\documentclass[11pt]{article}

\usepackage[final]{acl}

\usepackage{times}
\usepackage{latexsym}

\usepackage[T1]{fontenc}

\usepackage[utf8]{inputenc}

\usepackage{microtype}

\usepackage{inconsolata}

\usepackage{graphicx}

\usepackage{multirow}
\usepackage{enumitem}
\usepackage{amsmath}
\usepackage{amssymb}
\usepackage{booktabs}

\usepackage{booktabs}
\usepackage{multirow}
\usepackage{graphicx}
\usepackage{arydshln}
\usepackage[table]{xcolor}
\usepackage{caption}
\usepackage{algorithm}
\usepackage{algorithmic}

\definecolor{lavender}{RGB}{235,232,255}

\title{From Profiles to Steering Vectors: Global Sparse Priors and Local Semantic Calibration for Personalized Text Generation}

\author{
 \textbf{Liuji Chen\textsuperscript{1,2,}\thanks{Equal contribution.}},
 \textbf{Zeyu Zhang\textsuperscript{1,2,3,}\footnotemark[1]},
 \textbf{Xinyuan Zhang\textsuperscript{3}},
 \textbf{Shuai Nie\textsuperscript{3}},
\\
 \textbf{Qiang Liu\textsuperscript{1,2,}\thanks{Corresponding author.}},
 \textbf{Shu Wu\textsuperscript{1,2}},
 \textbf{Liang Wang\textsuperscript{1,2}},
\\
\\
 \textsuperscript{1}NLPR, Institute of Automation, Chinese Academy of Sciences, \\
 \textsuperscript{2}School of Artificial Intelligence, University of Chinese Academy of Sciences, \\
 \textsuperscript{3}HyperAI Team, Xiaomi Corporation
\\
\\
}

\begin{document}
\maketitle
\begin{abstract}
Personalized text generation requires models to capture user-specific writing styles from historical data. Existing approaches based on retrieval, parameter-efficient fine-tuning, or activation steering either introduce inference and storage overhead or struggle to separate stylistic signals from semantic content. We propose \textbf{\textit{GLASS}}, a training-free framework for personalized generation via \textbf{G}lobal–\textbf{L}ocal \textbf{A}ctivation \textbf{S}teering with \textbf{S}parse priors. GLASS uses sparse autoencoders to extract a global user-style prior from historical responses and constructs local contrastive style vectors over clustered interaction scenarios. During inference, it jointly injects global and local vectors into different model layers, enabling context-aware personalization without retrieval or parameter updates. Experiments on LaMP and LongLaMP show that GLASS outperforms retrieval-, fine-tuning-, and steering-based baselines across ROUGE metrics and LLM-as-judge evaluations. Further analyses show that SAE-based representations are more robust to topic and length shifts, suggesting better disentanglement of stylistic information from semantic residue. 
Our code is available at \url{https://github.com/zzFestinaLente/GLASS}.

\end{abstract}

\section{Introduction}

Large language models (LLMs) have demonstrated strong capabilities in generating fluent, coherent, and human-like text. As LLMs are increasingly deployed in user-facing writing scenarios, personalized text generation has become a crucial requirement. In applications such as email composition, review writing, dialogue assistance, and long-form content generation, users expect models not only to produce semantically appropriate responses, but also to reflect their individual writing habits, stylistic preferences, and communicative patterns \cite{zhao2025surveylargelanguagemodels, salemi2023lamp, kumar2024longlampbenchmarkpersonalizedlongform}.

Existing personalization methods can be broadly divided into three categories. Retrieval-based methods retrieve user-relevant historical texts and incorporate them into prompts to induce personalized generation through in-context learning \cite{gao2024retrievalaugmented, dong2024surveyincontextlearning}. Parameter-efficient fine-tuning (PEFT) methods adapt a subset of model parameters with user-specific data, allowing models to internalize personalized behaviors \cite{lin2024personasqpersonalizedsuggestedquestion, tan-etal-2024-personalized, zhuang2024hydramodelfactorizationframework}. More recently, activation steering methods provide a lightweight alternative by extracting style-related directions from user histories and intervening on model activations during inference \cite{silva-etal-2025-steering, du2025fintsefficientinferencetimepersonalization, arad-etal-2025-saes}.

Despite their effectiveness, retrieval- and fine-tuning-based approaches face practical limitations. Retrieval is typically driven by semantic similarity, which helps identify topically related examples but does not necessarily capture user-specific style. As a result, retrieved contexts may introduce noisy semantic content while providing weak evidence of writing preferences. Moreover, retrieval incurs additional inference latency and requires maintaining user histories. In contrast, PEFT-based methods require training and storing user-specific parameters, making them difficult to scale to large user populations \cite{zhang2025personalizedtextgenerationcontrastive}. Activation steering is therefore appealing, as it enables training-free personalization with low storage and inference overhead.

However, existing steering methods still struggle to separate stylistic signals from semantic content. Methods such as StyleVector \cite{zhang2025personalizedtextgenerationcontrastive} and Fints \cite{du2025fintsefficientinferencetimepersonalization} derive steering directions from hidden activations of historical examples, often through averaging or contrastive operations. Since user histories naturally contain both user-agnostic semantics and user-specific stylistic patterns, the resulting vectors may retain substantial semantic residue. This can cause the model to imitate topics or content from the history rather than capture the user's underlying style. Thus, a central challenge for activation-based personalization is how to disentangle style-relevant signals from semantic confounders in model representations.

\begin{figure}[t]
  \centering
  \includegraphics[width=\linewidth]{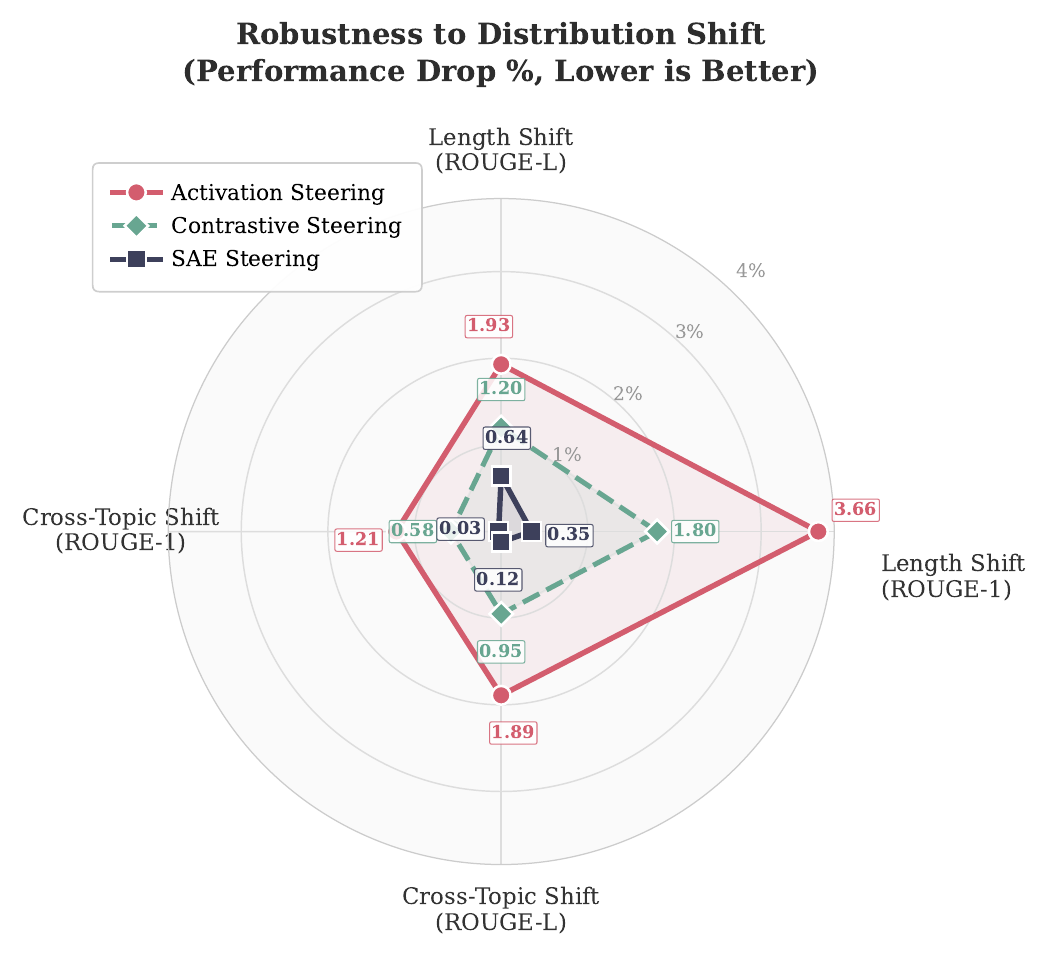}
  \caption{Performance degradation on downstream tasks when applying style features extracted by different methods after perturbing the length and topic distribution of user histories on LaMP-4.}
  \label{fig:robustness}
\end{figure}

Recent progress in sparse autoencoders (SAEs) \cite{zou2023transparency, liu2024context, rimsky-etal-2024-steering} offers a promising way to address this issue. SAEs project dense activations into a high-dimensional sparse feature space, where latent dimensions tend to capture more localized and interpretable features. We hypothesize that such sparse representations can better separate stylistic information from semantic content, leading to more reliable style features for personalization. To validate this intuition, we conduct a preliminary experiment on LaMP-4, where we perturb user histories along two dimensions: history length and dialogue topic. These perturbations create positive and negative sample pairs with substantial length discrepancies and topic shifts, respectively (Sec~\ref{sec:whyase}). As shown in Figure~\ref{fig:robustness}, SAE-based style features exhibit stronger robustness under both perturbations, suggesting that they reduce semantic residue more effectively while preserving user-specific stylistic information.

\begin{figure*}[t]
  \centering
  \includegraphics[width=\linewidth]{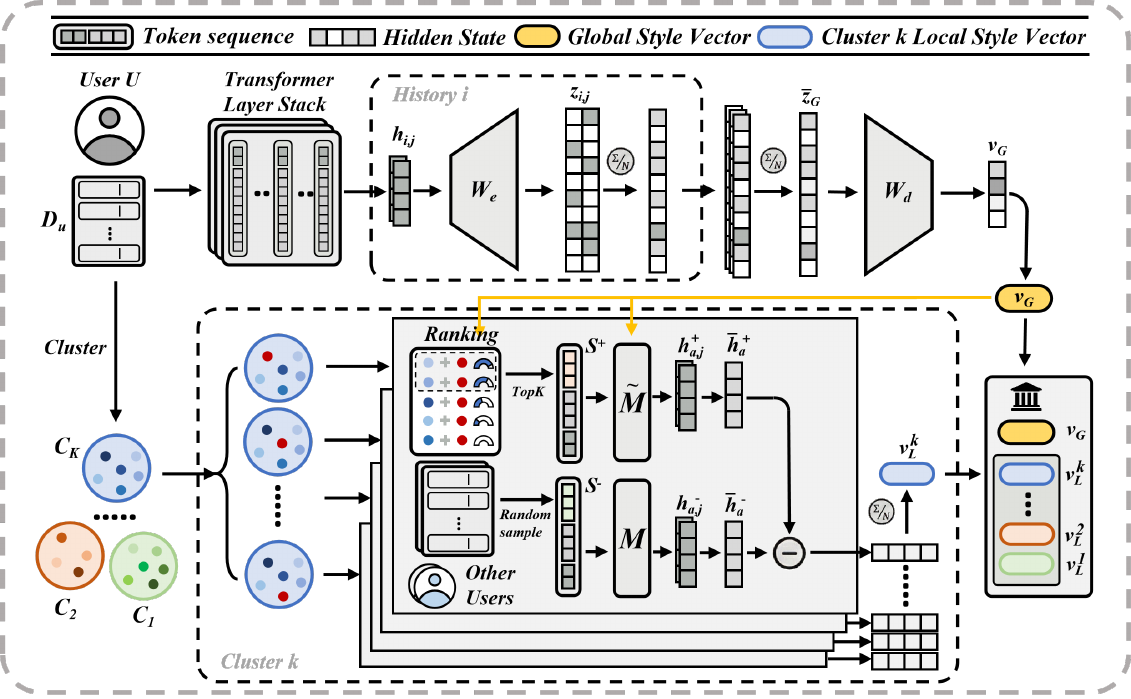}
  \caption{Overview of GLASS.}
  \label{fig:overview}
\end{figure*}

Motivated by this observation, we propose \textbf{GLASS} (\textbf{G}lobal--\textbf{L}ocal \textbf{A}ctivation \textbf{S}teering with \textbf{S}parse priors), a training-free personalization framework that performs activation steering in the SAE latent space. GLASS models user style at multiple granularities. It first extracts a global style representation from the user's historical data to capture overall writing tendencies. It then identifies local style patterns by clustering user histories and deriving cluster-specific steering directions, enabling context-dependent personalization. During inference, GLASS jointly applies global and local steering signals to guide generation while reducing semantic contamination from user histories. An overview of GLASS is shown in Figure~\ref{fig:overview}.

We evaluate GLASS on the \textbf{LaMP} \cite{salemi2023lamp} and \textbf{LongLaMP} \cite{kumar2024longlampbenchmarkpersonalizedlongform} benchmarks. Experimental results show that GLASS consistently outperforms retrieval-, fine-tuning-, and steering-based baselines while requiring no user-specific training and only minimal per-user storage. Our contributions are summarized as follows:
\begin{itemize}
    \item We identify semantic residue as a key limitation of existing activation-based personalization methods and show that SAE-based representations provide more robust style features under semantic and length perturbations.
    \item We propose \textbf{GLASS}, a training-free global--local activation steering framework that models both overall and context-dependent user writing styles in the SAE latent space.
    \item Extensive experiments on LaMP and LongLaMP demonstrate that GLASS achieves strong personalized generation performance with high efficiency and scalability.
\end{itemize}

\section{Preliminary}
Before introducing our proposed method, we first briefly present the problem formulation and the relevant background techniques utilized in our framework, namely Sparse Autoencoders (SAEs) and Activation Steering.

\subsection{Problem Formulation}
The goal of personalized text generation is to infer a user's writing style from their historical interaction data and to generate personalized text conditioned on both the user's style and a specific task requirement, such as email composition. Formally, for a given user $u$, let $D_u = \{(x_i, y_i)\}_{i=1}^{|D_u|}$ denote the user's historical interaction data, where $x_i$ represents a prompt specifying a particular task and $y_i$ denotes the corresponding text written in the user's authentic style. Given a new user request $x$, a model $M$ is expected to produce a personalized output $\hat{y} = M(x, D_u)$ that satisfies the generation requirements of user $u$ while preserving their stylistic nuances.

\subsection{Sparse Autoencoders}
Sparse Autoencoders (SAEs) have emerged as a powerful tool for interpreting the internal representations of Large Language Models (LLMs). LLMs typically represent concepts as dense vectors in a superposition state, making it difficult to isolate individual attributes such as writing style. SAEs address this by decomposing these dense activations into an overcomplete set of sparse, interpretable features.

Given an activation vector $h \in \mathbb{R}^{d_{model}}$ from a specific layer of an LLM, an SAE aims to reconstruct $h$ through a sparse intermediate representation. It consists of an encoder and a decoder. The encoder projects $h$ into a higher-dimensional feature space $f \in \mathbb{R}^{d_{sparse}}$ (where $d_{sparse} \gg d_{model}$), typically using an affine transformation followed by a non-linearity (e.g., ReLU) to enforce sparsity:
\begin{equation}
    f = \text{ReLU}(W_e h + b_e),
\end{equation}
where $W_e \in \mathbb{R}^{d_{sparse} \times d_{model}}$ and $b_e \in \mathbb{R}^{d_{sparse}}$ are the encoder weights and bias, respectively. The decoder then approximates the original activation:
\begin{equation}
    \hat{h} = W_d f + b_d,
\end{equation}
where $W_d$ and $b_d$ correspond to the decoder parameters. The training objective minimizes a combination of the reconstruction error (usually MSE) and an $L_1$ regularization penalty on $f$ to encourage sparsity:
\begin{equation}
    \mathcal{L} = ||h - \hat{h}||_2^2 + \lambda ||f||_1.
\end{equation}
In the context of our work, SAEs serve to disentangle stylistic features from the entangled activation space of the base model.

\subsection{Activation Steering}
Activation Steering refers to a lightweight inference-time intervention technique that controls an LLM's generation behavior by modifying its internal activations. Unlike fine-tuning, which updates model parameters, steering operates by injecting a "steering vector" that represents a desired concept or attribute into the forward pass.

Formally, let $h_l$ be the hidden state at layer $l$ of the model. To steer the model towards a specific attribute encoded by a vector $v_{steer}$, we modify the activation as follows:
\begin{equation}
    h'_l = h_l + \alpha \cdot v_{steer},
\end{equation}
where $\alpha$ is a scalar coefficient determining the injection strength. This intervention biases the model's probability distribution over the vocabulary, encouraging the generation of text that aligns with the target attribute (e.g., the user's specific writing style) while maintaining coherence with the input prompt.

\section{Method}
In this section, we introduce \textbf{GLASS}, which models user style to steer LLMs toward personalized generation at two granularities: a global representation for stable writing habits and local representations for context-dependent variations. GLASS first builds a style-vector bank offline, as illustrated in Figure~\ref{fig:overview}, and then retrieves and injects the corresponding vectors during inference, as shown in Figure~\ref{fig:infer}.

\subsection{Global Style Modeling via Sparse Priors}
The global style representation $v_G$ captures invariant stylistic user-level traits of a user $u$ such as lexical choice, sentence length, and tone.

Given a user history $D_u = \{(x_i, y_i)\}_{i=1}^{|D_u|}$, we feed each user-history example into the model under \emph{teacher forcing} and extract the hidden state $h_{i,j}$ at layer $L_G$ for the $j$-th token of response $y_i$. These dense activations are projected into the sparse feature space using the pretrained SAE encoder:
\begin{equation}
    z_{i,j} = \text{ReLU}(W_e h_{i,j} + b_e).
\end{equation}
Let $N_{token} = \sum_{i=1}^{|D_u|} |y_i|$ denote the total number of response tokens. We compute the global sparse profile by averaging sparse activations over all user responses:
\begin{equation}
    \bar{z}_G = \frac{1}{N_{token}} \sum_{i=1}^{|D_u|} \sum_{j=1}^{|y_i|} z_{i,j}.
\end{equation}
Finally, the global style vector $v_{G}$ is reconstructed by decoding this averaged sparse representation back into the residual stream space of the model:
\begin{equation}
    v_{G} = W_d \bar{z}_G + b_d.
\end{equation}
This token-level aggregation preserves recurring stylistic patterns while smoothing out instance-specific semantic variation.

\subsection{Local Style Modeling via Contrastive Clusters}
Although the global style vector $v_G$ summarizes the overall stylistic tendencies of a user, behavior is highly context-dependent, such as in formal emails versus casual forum posts. We therefore derive scenario-specific contrastive directions, denoted as local style vectors $v_C$.

\subsubsection{Scenario Clustering.}
We represent each prompt $x_i$ with TF-IDF features and apply K-means clustering to obtain $K$ clusters $\{C_1,\dots,C_K\}$, where $K$ is selected by maximizing the silhouette score. Each cluster approximates a recurring interaction scenario, and we extract a local style vector $v_C^{(k)}$ for each $C_k$.

\subsubsection{SAE-Primed Forward Pass.}
\label{sec:priming}
Local style extraction is challenging because activations can be dominated by content and generic generation priors, causing mined directions to capture semantics rather than style. To mitigate this, we define a primed model $\tilde{M}$ that injects $v_G$ into the residual stream at layer $L_G$ when extracting each $v_C^{(k)}$:
\begin{equation}
\tilde{h}_{L_G} = h_{L_G} + \gamma \cdot v_G, \end{equation} where $\gamma$ controls the injection strength. This priming biases subsequent ranking and contrastive extraction toward user style.

\subsubsection{Positive-Negative Contrastive Sample Construction.}

For each user $u$ and cluster $C_k$, we construct anchor-centered positive and negative samples for subsequent contrastive extraction. We first select an anchor set $\mathcal{A}_k \subset C_k$ and use each $(x_a,y_a)\in\mathcal{A}_k$ as a held-out target. The remaining examples form the candidate set $\mathcal{R}_k = C_k \setminus \mathcal{A}_k$, where $(x_m,y_m)\in\mathcal{R}_k$.

For positive sample construction, we evaluate each candidate in $\mathcal{R}_k$ as a one-shot demonstration for the anchor query $x_a$. We use ROUGE-L as the scoring function $\operatorname{score}(\cdot,\cdot)$. Let $p_m^a = \operatorname{Prompt}((x_m,y_m),x_a)$ be the prompted input; its utility score is
\begin{equation}
s_m^a = \operatorname{score}(\tilde{M}(p_m^a),\, y_a).
\end{equation}

After that, we sort the candidates in descending order and use the top-$n$ examples to form the positive $n$-shot context $S^+(x_a)=[(x_i,y_i)]_{i\in\mathcal{I}_a}$, where $\mathcal{I}_a$ denotes the indices of the selected candidates.

For negative sample construction, we sample $n$ histories from other users to form the negative $n$-shot context $S^-(x_a)=[(x_j,y_j)]_{j=1}^{n}$, where each $(x_j,y_j)$ is drawn from $\bigcup_{u'\neq u}D_{u'}$ without any cluster constraint.

To filter noisy pairs, we define the personalization score under positive and negative contexts as
\begin{equation}
\begin{aligned}
s^+_a &= \operatorname{score}(\tilde{M}([S^+(x_a);x_a]),\, y_a), \\
s^-_a &= \operatorname{score}(M([S^-(x_a);x_a]),\, y_a),
\end{aligned}
\end{equation}
and retain anchors satisfying $s^+_a > s^-_a$.

\begin{figure}[t]
  \centering
  \includegraphics[width=\linewidth]{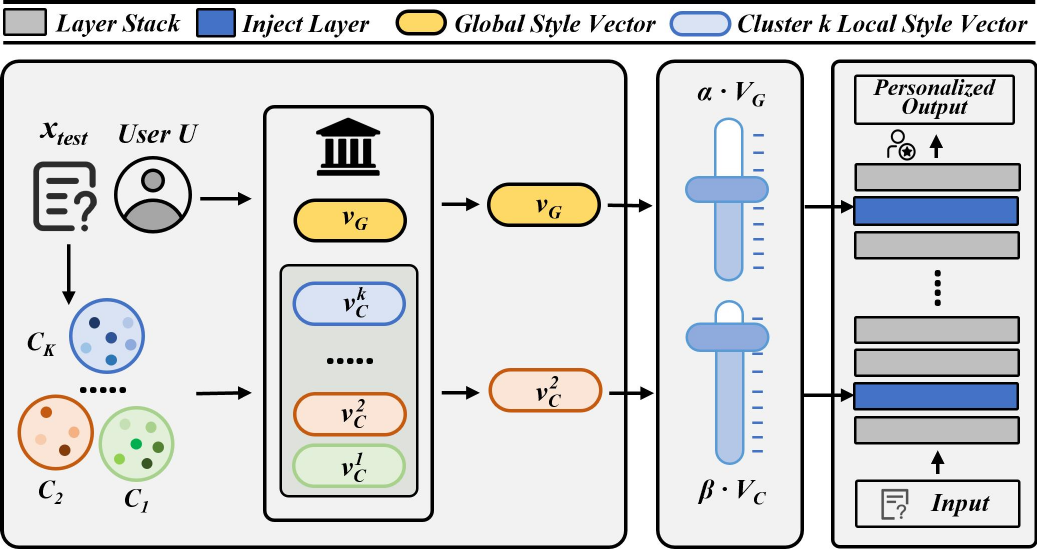}
  \caption{The inference procedure of GLASS.}
  \label{fig:infer}
\end{figure}

\subsubsection{Contrastive Activation Extraction.}
At target layer $L_C$, we contrast answer-token activations under positive and negative contexts.

For positives, we teacher-force $y_a$ through the primed model and extract the hidden state at layer $L_C$ for each answer token $t_{a,j}$ ($j=1,\dots,|y_a|$):
\begin{equation}
h^{+}_{a,j} = h_{L_C}\!\big(\tilde{M}([S^+(x_a); x_a; y_a])\big)\big|_{t_{a,j}}.
\end{equation}

For negatives, the base model first generates $\hat{y}^{-}_a=M([S^-(x_a); x_a])$, and we extract the hidden state for each generated token $\hat{t}_{a,j}$ ($j=1,\dots,|\hat{y}^{-}_a|$):
\begin{equation}
h^{-}_{a,j} = h_{L_C}\!\big(M([S^-(x_a); x_a; \hat{y}^{-}_a])\big)\big|_{\hat{t}_{a,j}}.
\end{equation}

We then apply mean pooling over answer tokens to obtain fixed-length representations:
\begin{equation}
\bar{h}^{+}_a = \frac{1}{|y_a|}\sum_{j=1}^{|y_a|} h^{+}_{a,j},
\qquad
\bar{h}^{-}_a = \frac{1}{|\hat{y}^{-}_a|}\sum_{j=1}^{|\hat{y}^{-}_a|} h^{-}_{a,j}.
\end{equation}
The cluster local style vector averages contrastive directions over anchors in $C_k$:
\begin{equation}
v_C^{(k)} = \mathbb{E}_{(x_a,y_a)\sim \mathcal{A}_k}\big[\bar{h}^{+}_a - \bar{h}^{-}_a\big].
\end{equation}

This subtraction removes shared semantic content and retains the local stylistic residual for $C_k$. All $v_C^{(k)}$ are precomputed and stored in a vector bank for inference-time steering.

\subsection{Adaptive Dual-Stream Steering}

During inference (Shown in Figure~\ref{fig:infer}), GLASS assigns a new request $x_{test}$ to the nearest cluster $C^*$, retrieves $v_C^*$ and the global style vector $v_G$, and injects them into the residual stream at different layers. The strengths of global and local steering are controlled by $\alpha$ and $\beta$:
\begin{align}
    \tilde{h}_{L_G} &= h_{L_G} + \alpha \cdot v_G, \\
    \tilde{h}_{L_C} &= h_{L_C} + \beta \cdot v_C^*.
\end{align}

This dual-stream intervention keeps $\hat{y}$ aligned with the user's overall style while adapting it to the current scenario.

\section{Experiments}
\begin{table*}[t]
  \centering

  \resizebox{\textwidth}{!}{%
  \begin{tabular}{@{}lcccccccccccc@{}}
    \toprule
    \multirow{2}{*}{Method} 
    & \multicolumn{3}{c}{\textbf{LaMP-4}} 
    & \multicolumn{3}{c}{\textbf{LaMP-5}} 
    & \multicolumn{3}{c}{\textbf{LongLaMP-3}} 
    & \multicolumn{3}{c}{\textbf{LongLaMP-4}} \\
    \cmidrule(lr){2-4} 
    \cmidrule(lr){5-7} 
    \cmidrule(lr){8-10} 
    \cmidrule(lr){11-13}
    & R-1 & R-L & Judge 
    & R-1 & R-L & Judge 
    & R-1 & R-L & Judge 
    & R-1 & R-L & Judge \\
    \midrule

    \multicolumn{13}{l}{\textit{\textbf{Non-personalized Method}}} \\
    Non-Pers. 
    & 0.1547 & 0.1311 & 1.12 
    & 0.4433 & 0.3880 & 1.13 
    & 0.1246 & 0.0733 & 1.10 
    & 0.1064 & 0.0728 & 1.08 \\

    \midrule

    \multicolumn{13}{l}{\textit{\textbf{Retrieval-based Methods}}} \\
    ICL 
    & 0.1183 & 0.1051 & 1.16
    & 0.1618 & 0.1377 & 1.27 
    & 0.1048 & 0.0605 & 1.16 
    & 0.1194 & 0.0752 & 1.20 \\
    RAG 
    & 0.1263 & 0.1126 & 1.29 
    & 0.2270 & 0.1935 & 1.47 
    & 0.1098 & 0.0658 & 1.31
    & \underline{0.1277} & 0.0786 & 1.49 \\

    \midrule

    \multicolumn{13}{l}{\textit{\textbf{PEFT-based Methods}}} \\
    SFT 
    & 0.1704 & 0.1432 & 1.86 
    & \underline{0.4621} & \underline{0.4079} & \underline{2.06} 
    & 0.1549 & 0.1022 & 1.78 
    & 0.1095 & 0.0774 & 1.45 \\
    OPPU 
    & \underline{0.1776} & 0.1511 & \underline{1.96} 
    & 0.4536 & 0.3897 & 2.01 
    & 0.1768 & \underline{0.1247} & 1.98 
    & \underline{0.1277} & \underline{0.0944} & \underline{1.92} \\

    \midrule

    \multicolumn{13}{l}{\textit{\textbf{Steering-based Methods}}} \\
    AS 
    & 0.1556 & 0.1360 & 1.77 
    & 0.4179 & 0.3621 & 1.97 
    & 0.1654 & 0.1097 & 1.87 
    & 0.1172 & 0.0835 & 1.75 \\
    StyleVector 
    & 0.1727 & \underline{0.1524} & 1.95 
    & 0.4510 & 0.3937 & 2.02 
    & \underline{0.1773} & 0.1205 & 1.98 
    & 0.1272 & 0.0876 & 1.85 \\
    Fints 
    & 0.1677 & 0.1438 & 1.85 
    & 0.4452 & 0.3822 & 2.04 
    & \textbf{0.2034} & 0.1212 & \underline{2.04} 
    & 0.1118 & 0.0763 & 1.50 \\

    \midrule

    \rowcolor{lavender!25}
    GLASS 
    & \textbf{0.1872} & \textbf{0.1640} & \textbf{2.09} 
    & \textbf{0.4875} & \textbf{0.4300} & \textbf{2.16} 
    & 0.1742 & \textbf{0.1288} & \textbf{2.13} 
    & \textbf{0.1284} & \textbf{0.0947} & \textbf{1.99} \\

    \bottomrule
  \end{tabular}%
  }
\caption{Performance comparison of different methods on LaMP and LongLaMP benchmarks. The best results are indicated in bold, and the second-best results are \underline{underlined}.}
  \label{tab:main_res}
\end{table*}

\subsection{Experiment Setup}
\subsubsection{Datasets and Evaluation}
We evaluate GLASS on two widely used personalized text generation benchmarks, LaMP \cite{salemi2023lamp} and LongLaMP \cite{kumar2024longlampbenchmarkpersonalizedlongform}, covering short- and long-term personalization. We exclude the email subset for privacy reasons and omit LaMP-7 due to its fully synthetic LLM-generated distribution. Accordingly, we use LaMP-4, LaMP-5, LongLaMP-3, and LongLaMP-4. Following prior work \cite{du2025fintsefficientinferencetimepersonalization, zhuang2024hydramodelfactorizationframework}, we sample 200 users appearing in both training and test sets. We report ROUGE-1 and ROUGE-L as primary metrics, following previous studies \cite{zhang2025personalizedtextgenerationcontrastive, du2025fintsefficientinferencetimepersonalization, zhang2025primelargelanguagemodel}. Since ROUGE mainly captures lexical and structural overlap, we also conduct an \textbf{LLM-as-judge} evaluation using GPT-5 across five personalization criteria: Style Consistency, Preference/Persona Alignment, Content Personalization, Distinctiveness, and Overall Personalization. We report the average score, with details in Appendix~\ref{app:llmasjudge}.
\subsubsection{Baselines}
To evaluate the effectiveness of our method, we compare it with strong baselines from three categories. (1) \textbf{Retrieval-based methods.} We evaluate in-context learning (ICL) by randomly sampling \(k\) instances from each user's history to construct a \(k\)-shot personalized prompt. We also include representative retrieval-augmented generation (RAG) methods for personalization. (2) \textbf{PEFT-based methods.} We compare with supervised fine-tuning (SFT) and OPPU \cite{tan-etal-2024-democratizing}, a personalized parameter-efficient fine-tuning method. (3) \textbf{Steering-based methods.} We consider three activation steering baselines: AS, which uses the average activation over all user histories as the steering vector; StyleVector \cite{zhang2025personalizedtextgenerationcontrastive}, which contrasts activations between personalized and non-personalized texts; and Fints \cite{du2025fintsefficientinferencetimepersonalization}, which combines retrieval with vector-based steering.

\subsubsection{Implementation Details}
All experiments use LLaMA-3-8B \cite{grattafiori2024llama3herdmodels} as the backbone. By default, GLASS adopts an EleutherAI pretrained SAE with a 32$\times$ expansion factor. We also evaluate alternative SAE models and report the results in Appendix~\ref{app:sensitivity}. For all methods that require retrieval, we use Contriever \cite{izacard2021contriever} for all retrieval-based methods. To ensure fair comparison, all methods use the same history size ($k=5$), including ICL demonstrations, retrieved histories, and positive--negative pairs for Fints \cite{du2025fintsefficientinferencetimepersonalization} and GLASS. We select task-specific intervention layers and strengths $(\alpha,\beta,\gamma)$ on validation data, with full implementation details reported in Appendix~\ref{app:implementation}.

\subsection{Research Questions}
We evaluate GLASS starting from the research questions below.

\begin{itemize}[leftmargin=7pt,nosep]
\item \textbf{RQ1}: Does GLASS outperform the competing baselines?
\item \textbf{RQ2}: Does each component in GLASS contribute uniquely to personalization performance?
\item \textbf{RQ3}: Does semantic redundancy degrade personalization? Can SAEs effectively disentangle stylistic from semantic information?
\item \textbf{RQ4}: Does GLASS incur additional inference latency?
\item \textbf{RQ5}: How do the choice of steering layers and the intervention strength influence the performance of GLASS?
\item \textbf{RQ6}: How sensitive is GLASS to key design choices such as clustering strategy, historical size, SAE model selection and feature aggregation?
\end{itemize}

\subsection{Main Results (RQ1)}
\label{exp:main_res}
As shown in Table~\ref{tab:main_res}, GLASS consistently achieves the strongest overall performance across LaMP and LongLaMP benchmarks. On LaMP-4 and LaMP-5, GLASS obtains the best results on all three metrics, including both lexical-overlap metrics (R-1 and R-L) and the LLM-as-judge personalization score (Detailed in Appendix~\ref{app:llmasjudge}). Compared with the strongest competing baselines, GLASS improves R-1/R-L/Judge from 0.1776/0.1524/1.96 to 0.1872/0.1640/2.09 on LaMP-4, and from 0.4621/0.4079/2.06 to 0.4875/0.4300/2.16 on LaMP-5. These gains suggest that GLASS not only better matches the surface form of personalized outputs, but also produces responses that are judged as more aligned with user-specific writing preferences.

The advantage of GLASS remains evident on the more challenging LongLaMP setting, where each user contains longer and more diverse historical contexts. On LongLaMP-3, GLASS achieves the best R-L and Judge scores, while ranking fourth in terms of R-1. On LongLaMP-4, GLASS again achieves the best performance across all metrics. Overall, GLASS ranks first on 11 out of 12 evaluation metrics and fourth on the remaining one, demonstrating robust personalization ability across both short-history and long-history scenarios. These results validate the effectiveness of combining global user-level style information with local cluster-specific style representations for personalized generation.

\subsection{Ablation Study (RQ2)}
\label{exp:ablation}

\subsubsection{Effect of Global Sparse Prior and Local Style Vectors}

\begin{table}[t]
  \centering
  
  \resizebox{\columnwidth}{!}{%
  \begin{tabular}{@{}lcccc@{}}
    \toprule
     & \multicolumn{2}{c}{\textbf{LaMP-4}} & \multicolumn{2}{c}{\textbf{LaMP-5}} \\
    \cmidrule(lr){2-3} \cmidrule(lr){4-5}
    & ROUGE-1 & ROUGE-L & ROUGE-1 & ROUGE-L   \\
    \midrule
    GLASS & 0.1872 & 0.1640& 0.4875 & 0.4300 \\
    w/o \(v_G\) & 0.1624 & 0.1379 & 0.4447 & 0.3895 \\
    w/o \(v_C\) & 0.1772 & 0.1552& 0.4670 & 0.4099 \\
    \bottomrule
  \end{tabular}%
  }
  \caption{Ablation study on the contribution of global sparse prior ($v_G$) and local contrastive style vector ($v_C$).}
  \label{tab:ablation}
\end{table}

In this section, we conduct ablation experiments to validate the effectiveness of the two core components of \textbf{GLASS}. The experimental results are reported in Table~\ref{tab:ablation}. The results indicate that removing the global style component causes the most significant performance degradation, yielding results that are close to those of Fints and only marginally better than the non-personalized baseline. In addition, removing the local style component also leads to a clear drop in performance, demonstrating that the two modules are complementary rather than redundant. 

\subsubsection{Effect of Contrastive Extraction}

\begin{figure}[t]
  \centering
  \includegraphics[width=\linewidth]{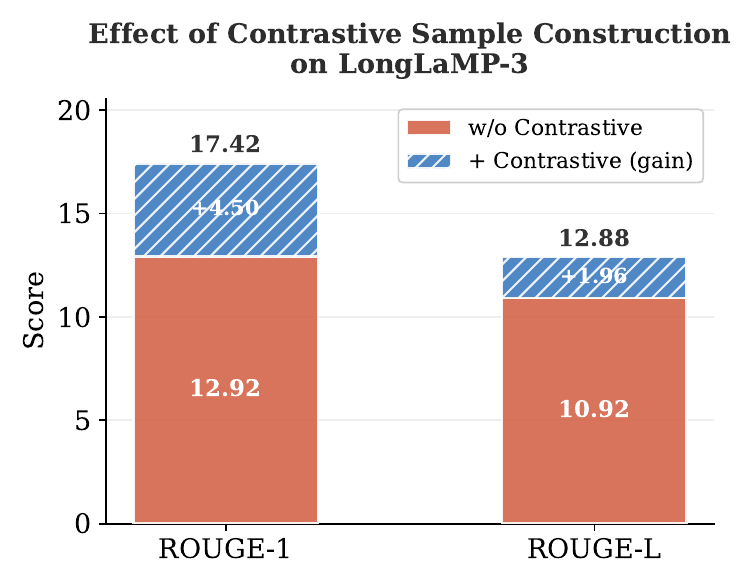}
  \caption{Impact of contrastive sample construction on LongLaMP-3. Introducing contrastive samples substantially improves both ROUGE-1 and ROUGE-L, demonstrating the effectiveness of contrastive activation extraction for capturing user-specific style features.}
  \label{fig:contrastive}
\end{figure}

We ablate the contrastive mechanism by removing the negative sample branch and constructing the local style vector solely from positive activations. As shown in Figure~\ref{fig:contrastive}, this variant leads to a substantial performance degradation across both metrics, demonstrating that contrastive extraction is essential for suppressing content-correlated signals and isolating genuine stylistic directions.

\subsection{Analysis}
\subsubsection{Effect of Semantic Redundancy and SAE Disentanglement (RQ3)}
\label{sec:whyase}
As discussed in the Introduction, the central idea of GLASS is that SAE can better disentangle semantic information from stylistic information in user histories, thereby extracting more reliable style representations. We provide further evidence for this claim here. First, the results in Figure~\ref{fig:robustness} show that SAE-based steering is more robust than conventional methods when the length and topic of user histories are perturbed. This suggests that the extracted style features contain less residual semantic information, even under substantial distributional shifts between positive and negative samples. Second, the main results in Table~\ref{tab:main_res} show that GLASS achieves strong performance across multiple downstream personalization tasks, further demonstrating the effectiveness of SAE. Finally, the ablation results in Table~\ref{tab:ablation} show that removing SAE leads to the largest performance drop, indicating that the style features extracted by SAE are particularly critical to GLASS. Taken together, these findings support our claim that SAE can more effectively capture user-specific stylistic features.

Together, the main results and the robustness analysis jointly confirm that semantic disentanglement is critical for effective personalization, and that SAEs provide a principled mechanism to achieve it.

\subsubsection{Efficiency Analysis (RQ4)}

\begin{figure}[t]
  \centering
  \includegraphics[width=\linewidth]{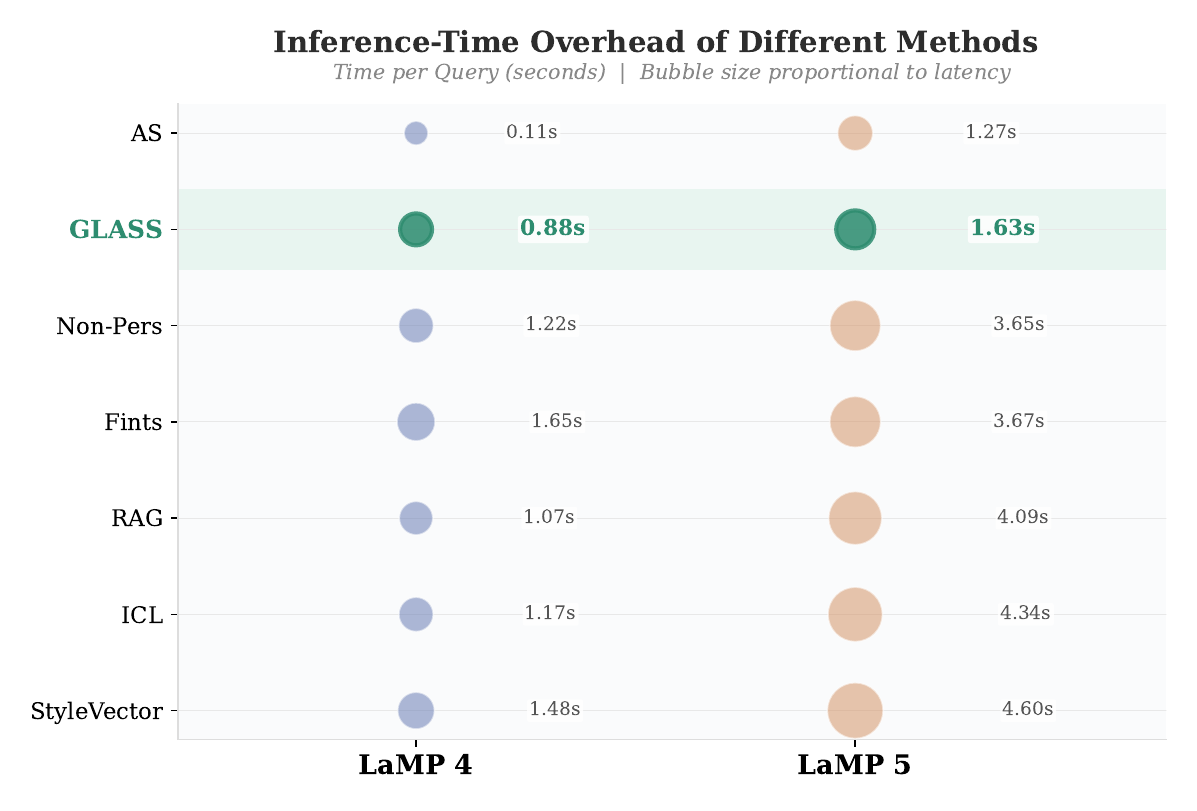}
  \caption{Inference-time overhead of different methods.}
  \label{fig:time}
\end{figure}

 Figure~\ref{fig:time} compares the inference latency of GLASS with all baselines on LaMP-4 and LaMP-5. The key observation is that GLASS achieves a strong efficiency--performance trade-off. Unlike ICL and RAG-based methods, GLASS avoids prompt expansion and retrieval overhead, introducing only lightweight representation-level steering at inference time. We also compare offline costs with personalized PEFT methods in Appendix~\ref{app:offline}. As a training-free method, GLASS requires much lower construction time and per-user storage than LoRA-based personalization, while maintaining stable style vectors over time.

\subsubsection{Analysis of Intervention Strengths and Layers (RQ5)}

Our experimental results are shown in Figures~\ref{fig:layer_strength},~\ref{fig:pairwise_vg_vc}, and~\ref{fig:pairwise_vg_sp}. We summarize the key findings here and leave detailed analyses to Appendix~\ref{app:layer_strength_fig} and Appendix~\ref{app:pairwise_fig}. Overall, GLASS benefits most from depth-aware and moderately scaled steering. The global vector $v_G$ is more effective for upper-layer user-level style modulation, while the local vector $v_C$ complements it at intermediate layers by refining context-specific stylistic cues. In addition, the SAE-primed forward pass ($SP$) mainly improves offline local vector extraction rather than acting as an independent inference-time signal. These results suggest that effective personalization depends on balancing global and local style interventions across suitable model depths.

\subsubsection{Sensitivity Analysis (RQ6)}
\label{sec:sensitivity}

In this section, we discuss the sensitivity of GLASS to several key design choices and provide a detailed discussion of these choices in the Appendix~\ref{app:appex}.

\textbf{(1) Clustering strategy}. As shown in Table~\ref{tab:app_cluster}, adaptive clustering, which selects $K$ using the silhouette score, consistently outperforms fixed cluster assignment. This confirms that dynamically identifying interaction scenarios enables more precise local style modeling.

\textbf{(2) Aggregation strategy}. Table~\ref{tab:app_sae} shows that mean aggregation achieves the best performance among the four strategies for constructing $v_G$. In contrast, max pooling and last-token pooling are largely ineffective, suggesting that averaging token-level sparse activations provides a more reliable estimate of users’ distributional stylistic preferences.

\textbf{(3) Historical data sizes}. GLASS maintains strong stability across varying historical data sizes ($k \in \{1, 3, 5, 8, 10\}$), with $k=5$ providing the best trade-off between performance and efficiency, shown in Table~\ref{tab:app_data_size}.

\textbf{(4) Different SAE models}. As shown in Table~\ref{tab:app_sae_model}, switching across SAE models with 64$\times$ expansion produces comparable results, while reducing to 8$\times$ expansion leads to moderate degradation, indicating that sufficient latent dimensionality is necessary for effective stylistic disentanglement.

\begin{table}[h]
  \centering
  \resizebox{\columnwidth}{!}{%
  \begin{tabular}{@{}lcc@{}}
    \toprule
    \textbf{SAE Model} & \textbf{LongLaMP-3} & \textbf{LongLaMP-4} \\
    \midrule
    Base (No SAE)         & 0.1148 & 0.1056 \\
    EleutherAI (64$\times$) & 0.1778 & 0.1194 \\
    LlamaScope (8$\times$)  & 0.1467 & 0.1110 \\
    THUKEG (64$\times$)     & 0.1771 & 0.1158 \\
    \bottomrule
  \end{tabular}%
  }
  \caption{Performance of GLASS with different SAE models (R-L).}
  \label{tab:app_sae_model}
\end{table}

\subsection{Case Study}
To intuitively demonstrate the personalized generation quality of GLASS, we provide detailed discussions and several case studies in Appendix~\ref{app:case_study}. We also use an LLM-as-judge evaluation to analyze how well different methods align with users’ personalized preferences.

\section{Conclusion}
We introduced GLASS, a training-free personalization method that steers LLMs with sparse global and local style representations. By leveraging sparse autoencoders, GLASS reduces semantic residue in activation-based personalization and captures both stable user-level writing patterns and scenario-specific stylistic variations. Experiments on LaMP and LongLaMP demonstrate that GLASS consistently improves personalized generation over retrieval-, PEFT-, and steering-based baselines while maintaining low inference overhead. Ablation and robustness analyses further confirm the importance of sparse global priors, local contrastive vectors, and SAE-based disentanglement. These findings suggest that sparse representation-level steering is a scalable and effective paradigm for personalized text generation.

\section*{Limitations}

While GLASS demonstrates strong effectiveness and efficiency for personalized text generation, several directions remain open for future work. First, GLASS derives user style vectors in a training-free manner, making the framework lightweight and scalable. However, the current vector construction process still relies on fixed SAE representations and activation-level aggregation. Although sparse representations help reduce semantic residue, future work could explore lightweight refinement strategies to further disentangle stylistic signals from content-related factors while preserving the storage and inference efficiency of GLASS.

Second, our evaluation focuses on standard personalized text generation benchmarks, including LaMP and LongLaMP. These benchmarks provide controlled and reproducible settings for comparing personalization methods, but real-world personalization often involves more diverse and dynamic writing contexts. For example, the same user may adopt different stylistic registers across formal emails, academic writing, reviews, and casual social media posts. Future benchmarks could better assess whether personalization methods can preserve task-appropriate stylistic variation while avoiding interference among conflicting domain-specific patterns.

\section*{Ethical Considerations}
In this work, all training data and frameworks used by GLASS comply with their respective licenses and do not infringe on any privacy; they are used solely for academic research purposes.

\section*{Acknowledgements}
This work is supported by National Natural Science Foundation of China (62372454, 62576339, 92570204). 

\bibliography{custom,references}

\appendix

\section{Related Work}
\subsection{Personalized Text Generation}
The rapid advancement of large language models (LLMs) has fundamentally reshaped content generation paradigms, shifting the field from uniform, generic outputs toward increasingly sophisticated forms of personalized text generation. Existing personalization techniques can be broadly categorized into three methodological paradigms.

The first paradigm consists of Retrieval-Augmented Generation (RAG)–based methods, which leverage users' historical content through dynamic retrieval mechanisms. Early studies~\cite{zhang-etal-2024-llm-based, Salemi, Salemib} established foundational retrieval-based personalization frameworks, while subsequent work has substantially extended these approaches. In particular, profile-augmented prompting strategies have been proposed to more effectively incorporate user information into the generation process~\cite{richardson2023integratingsummarizationretrievalenhanced, zhang-2024-guided, tan2025largelanguagemodelsunderstand}. In addition, feedback-driven retrieval optimization has been shown to further enhance personalization fidelity by iteratively refining retrieved content~\cite{Salemib}.

The second paradigm is based on Parameter-Efficient Fine-Tuning (PEFT), which enables personalization by adapting a small set of user-specific parameters via lightweight adapter modules. Empirical studies~\cite{Salemi} demonstrate that PEFT-based approaches—particularly those employing per-user adapter tuning~\cite{tan-etal-2024-personalized, zhuang2024hydramodelfactorizationframework, liu-etal-2025-llms, Ding2025kdd}—can achieve strong personalization performance while maintaining favorable computational efficiency.

The third paradigm centers on \emph{activation steering}. A key advantage of this line of work is that it is inherently \emph{training-free}: model parameters remain fixed, and personalization is realized by storing only a small number of user-specific steering vectors. Representative methods include StyleVector~\cite{zhang2025personalizedtextgenerationcontrastive}, which derives personalization signals through user-specific \emph{contrastive activations}, and Fints~\cite{du2025fintsefficientinferencetimepersonalization}, which extracts contrastive activations by retrieving semantically similar historical instances and exploiting their corresponding internal representations.

\subsection{Activation Steering}
Recent advances in activation engineering suggest that large language models represent high-level semantic concepts as approximately linear subspaces in their hidden activations \cite{zou2023transparency, liu2024context, rimsky-etal-2024-steering}. This geometric perspective makes it possible to steer model behavior at inference time via simple linear interventions on internal representations. Early work by Turner et al.  Alexander \cite{turner2024steeringlanguagemodelsactivation} introduced activation addition based on contrastively constructed steering vectors to control attributes such as sentiment and topic. Building on this, Rimsky et al. \cite{rimsky-etal-2024-steering} improved steering accuracy by leveraging mass-mean activation differences. In parallel, Zhang et al. \cite{zhang-etal-2024-truthx} used linear probes to identify truth-correlated attention heads and enhanced factuality through targeted modulation, while Chen et al. \cite{chen2024truthforestmultiscaletruthfulness} proposed multi-directional orthogonal steering to further strengthen truthfulness in generated outputs.

\subsection{Sparse Autoencoders}
Sparse Autoencoders (SAEs) have emerged as a prominent unsupervised method for interpreting the internal states of Large Language Models (LLMs). Addressing the challenge of polysemanticity—where individual neurons respond to multiple unrelated concepts due to superposition—SAEs project dense model activations into a significantly higher-dimensional, sparse feature space. By optimizing for both reconstruction fidelity and sparsity, SAEs effectively disentangle complex neural activity into interpretable, monosemantic features. Recent research has further refined this approach through architectural variants, such as Gated SAEs and TopK activation schemes, which aim to improve the trade-off between reconstruction accuracy and feature sparsity while mitigating training issues like dead latents \cite{arad-etal-2025-saes, shu-etal-2025-survey}.

\section{Algorithmic Description of GLASS}
\label{app:algorithm}

For completeness and reproducibility, we provide the full pseudocode of GLASS. Algorithm~\ref{alg:offline} describes the offline style-vector bank construction, corresponding to the \emph{Global Style Modeling via Sparse Priors} and \emph{Local Style Modeling via Contrastive Clusters} components of the method. Algorithm~\ref{alg:inference} describes the inference-time adaptive dual-stream steering procedure.

\begin{algorithm*}[t]
\caption{GLASS Offline Style Vector Bank Construction}
\label{alg:offline}
\small
\begin{algorithmic}[1]
\REQUIRE User history $\mathcal{D}_u = \{(x_i, y_i)\}_{i=1}^{|\mathcal{D}_u|}$; base model $M$; pretrained SAE encoder $(W_e, b_e)$ and decoder $(W_d, b_d)$; global layer $L_G$; local layer $L_C$; priming strength $\gamma$; number of positive examples $n$
\ENSURE Global style vector $v_G$; local style vector bank $\{v_C^{(k)}\}_{k=1}^{K}$

\STATE \textbf{// Phase 1: Global Style Modeling (Sec.~3.1)}
\FOR{each $(x_i, y_i) \in \mathcal{D}_u$}
    \FOR{each token $t_{i,j}$ in $y_i$}
        \STATE Extract $h_{i,j} \leftarrow h^{L_G}(M([x_i; y_i]))_{t_{i,j}}$ via teacher forcing
        \STATE $z_{i,j} \leftarrow \text{ReLU}(W_e \cdot h_{i,j} + b_e)$ \hfill $\triangleright$ Eq.~(5)
    \ENDFOR
\ENDFOR
\STATE $N_{\text{token}} \leftarrow \sum_{i=1}^{|\mathcal{D}_u|} |y_i|$
\STATE $\bar{z}_G \leftarrow \frac{1}{N_{\text{token}}} \sum_{i=1}^{|\mathcal{D}_u|} \sum_{j=1}^{|y_i|} z_{i,j}$ \hfill $\triangleright$ Eq.~(6)
\STATE $v_G \leftarrow W_d \cdot \bar{z}_G + b_d$ \hfill $\triangleright$ Eq.~(7)

\STATE \textbf{// Phase 2: Local Style Modeling (Sec.~3.2)}
\STATE Compute TF-IDF representations for all $\{x_i\}_{i=1}^{|\mathcal{D}_u|}$
\STATE Apply K-means clustering $\rightarrow \{C_1, \ldots, C_K\}$, selecting $K$ via silhouette score
\STATE Define primed model $\tilde{M}$: $\tilde{h}^{L_G} = h^{L_G} + \gamma \cdot v_G$ \hfill $\triangleright$ Eq.~(8)
\FOR{each cluster $C_k$, $k = 1, \ldots, K$}
    \STATE Select anchor set $\mathcal{A}_k \subset C_k$; candidate set $\mathcal{R}_k \leftarrow C_k \setminus \mathcal{A}_k$
    \FOR{each anchor $(x_a, y_a) \in \mathcal{A}_k$}
        \STATE \textbf{// Positive-Negative Construction (Sec.~3.2.3)}
        \FOR{each $(x_m, y_m) \in \mathcal{R}_k$}
            \STATE $s_m^a \leftarrow \text{score}(\tilde{M}(\text{Prompt}((x_m, y_m), x_a)),\; y_a)$ \hfill $\triangleright$ Eq.~(9)
        \ENDFOR
        \STATE $S^+(x_a) \leftarrow$ top-$n$ candidates by $s_m^a$ in descending order
        \STATE $S^-(x_a) \leftarrow$ sample $n$ histories from $\bigcup_{u' \neq u} \mathcal{D}_{u'}$
        \STATE $s_a^+ \leftarrow \text{score}(\tilde{M}([S^+(x_a); x_a]),\; y_a)$ \hfill $\triangleright$ Eq.~(10)
        \STATE $s_a^- \leftarrow \text{score}(M([S^-(x_a); x_a]),\; y_a)$
        \IF{$s_a^+ \leq s_a^-$}
            \STATE Discard anchor $(x_a, y_a)$; \textbf{continue}
        \ENDIF
        \STATE \textbf{// Contrastive Activation Extraction (Sec.~3.2.4)}
        \FOR{each answer token $t_{a,j}$ in $y_a$}
            \STATE $h_{a,j}^+ \leftarrow h^{L_C}(\tilde{M}([S^+(x_a); x_a; y_a]))_{t_{a,j}}$ \hfill $\triangleright$ Eq.~(11)
        \ENDFOR
        \STATE $\hat{y}_a^- \leftarrow M([S^-(x_a); x_a])$ \hfill $\triangleright$ Generate under negative context
        \FOR{each generated token $\hat{t}_{a,j}$ in $\hat{y}_a^-$}
            \STATE $h_{a,j}^- \leftarrow h^{L_C}(M([S^-(x_a); x_a; \hat{y}_a^-]))_{\hat{t}_{a,j}}$ \hfill $\triangleright$ Eq.~(12)
        \ENDFOR
        \STATE $\bar{h}_a^+ \leftarrow \frac{1}{|y_a|} \sum_{j=1}^{|y_a|} h_{a,j}^+$;\; $\bar{h}_a^- \leftarrow \frac{1}{|\hat{y}_a^-|} \sum_{j=1}^{|\hat{y}_a^-|} h_{a,j}^-$ \hfill $\triangleright$ Eq.~(13)
    \ENDFOR
    \STATE $v_C^{(k)} \leftarrow \mathbb{E}_{(x_a, y_a) \sim \mathcal{A}_k} \left[ \bar{h}_a^+ - \bar{h}_a^- \right]$ \hfill $\triangleright$ Eq.~(14)
\ENDFOR
\RETURN $v_G$, $\{v_C^{(k)}\}_{k=1}^{K}$
\end{algorithmic}
\end{algorithm*}

\begin{algorithm*}[t]
\caption{GLASS Inference: Adaptive Dual-Stream Steering}
\label{alg:inference}
\small
\begin{algorithmic}[1]
\REQUIRE Test request $x_{\text{test}}$; base model $M$; global style vector $v_G$; local style vector bank $\{v_C^{(k)}\}_{k=1}^{K}$ with cluster centroids $\{\mu_k\}_{k=1}^{K}$; global layer $L_G$; local layer $L_C$; steering strengths $\alpha$, $\beta$
\ENSURE Personalized output $\hat{y}$

\STATE Compute TF-IDF representation of $x_{\text{test}}$
\STATE $C^* \leftarrow \arg\min_{k} \| \text{TF-IDF}(x_{\text{test}}) - \mu_k \|$ \hfill $\triangleright$ Nearest cluster
\STATE Retrieve local style vector $v_C^* \leftarrow v_C^{(C^*)}$
\FOR{each decoding step}
    \STATE $\tilde{h}^{L_G} \leftarrow h^{L_G} + \alpha \cdot v_G$ \hfill $\triangleright$ Eq.~(15), global steering
    \STATE $\tilde{h}^{L_C} \leftarrow h^{L_C} + \beta \cdot v_C^*$ \hfill $\triangleright$ Eq.~(16), local steering
    \STATE Continue forward pass with modified activations
    \STATE Sample next token from output distribution
\ENDFOR
\RETURN $\hat{y}$
\end{algorithmic}
\end{algorithm*}

\section{Implementation Details}
\label{app:implementation}

All experiments use LLaMA-3-8B as the backbone model, including the base and instruction-tuned variants when required by the corresponding baseline. Models are loaded in \texttt{bfloat16}; inference is performed on a single GPU with left padding (\texttt{padding\_side=left}) and deterministic decoding (\texttt{do\_sample=False}). The full experimental pipeline is run on two NVIDIA A100 GPUs. Table~\ref{tab:app_impl_config} summarizes the task-specific generation lengths and GLASS intervention settings. Here, SAE denotes the global sparse prior $v_G$, Cluster denotes the local contrastive vector $v_C$, and SP denotes the SAE-primed forward pass used during local vector construction.

\begin{table*}[t]
  \centering
  \small
  \setlength{\tabcolsep}{4pt}
  \begin{tabular}{@{}lccccccc@{}}
    \toprule
    \textbf{Dataset} & \textbf{Max Tokens} & \textbf{Cluster Layer} & \textbf{Cluster Strength} & \textbf{SAE Layer} & \textbf{SAE Strength} & \textbf{SP Layer} & \textbf{SP Strength} \\
    \midrule
    LaMP-4       & 100 & 23 & 0.2 & 19 & 0.8 & 23 & 0.4 \\
    LaMP-5       & 150 & 23 & 0.6 & 19 & 0.4 & 23 & 1.0 \\
    LongLaMP-3   & 450 & 23 & 0.8 & 19 & 0.4 & 23 & 0.1 \\
    LongLaMP-4   & 400 & 23 & 0.7 & 19 & 0.4 & 23 & 0.1 \\
    \bottomrule
  \end{tabular}
  \caption{Task-specific generation limits and GLASS intervention configurations. Cluster corresponds to $v_C$, SAE corresponds to $v_G$, and SP corresponds to the SAE-primed forward pass.}
  \label{tab:app_impl_config}
\end{table*}

\section{Detailed Experiments}
\label{app:appex}
\subsection{LLM-as-Judge}
\label{app:llmasjudge}

ROUGE-1 and ROUGE-L mainly measure lexical and structural overlap, which only partially reflects whether a generation captures the target user's tone, preferences, and writing habits. We therefore use \textbf{GPT-5} as a rubric-based judge to assess personalization quality along five dimensions.

\paragraph{Evaluation dimensions.}
Each dimension is rated on a 1--5 Likert scale, where 1 indicates generic or unrelated output and 5 indicates strong alignment with the target user. The dimensions are:
\begin{itemize}[leftmargin=12pt,nosep]
    \item \textbf{Style Consistency (SC).} Match in tone, syntax, word choice, and formality.
    \item \textbf{Preference \& Persona Alignment (PP).} Alignment with the user's recurring preferences, organization, and voice.
    \item \textbf{Content Personalization (CP).} Alignment in content selection, emphasis, and framing.
    \item \textbf{Distinctiveness (D).} Departure from generic, user-agnostic text.
    \item \textbf{Overall Personalization (OP).} Overall likelihood that the text was authored by the target user.
\end{itemize}

\paragraph{Evaluation protocol.}
For each test instance, the judge receives the task instruction, sampled user-history pairs from $\mathcal{D}_u$, the test input $x_{\text{test}}$, the user-written reference, and one candidate output. It assigns an integer score in $\{1,2,3,4,5\}$ for each dimension with a short justification. All methods are evaluated with identical user contexts and prompt formatting; method names are anonymized and candidate order is randomized to reduce identity and position bias. We use deterministic decoding (temperature $=0$) and the same rubric prompt across all datasets; the full prompt is provided in Table~\ref{tab:judge_prompt}. For each method and dataset, we average scores over the same 200-user evaluation pool used in the main experiments and report both dimension-level scores and their arithmetic mean as \emph{Average}.

\paragraph{Results.}
Tables~\ref{tab:gpt5_judge_lamp4},~\ref{tab:gpt5_judge_lamp5},~\ref{tab:gpt5_judge_longlamp3}, and~\ref{tab:gpt5_judge_longlamp4} report the judge scores on LaMP-4, LaMP-5, LongLaMP-3, and LongLaMP-4, respectively. GLASS achieves the highest \emph{Average} and OP scores on all four datasets, while remaining at or near the top on the other dimensions. Its gains on SC and D suggest that global--local steering improves user-attributable style beyond what lexical-overlap metrics capture.

\begin{table}[t]
\centering
\small
\setlength{\tabcolsep}{4pt}
\resizebox{\columnwidth}{!}{
\begin{tabular}{lcccccc}
\toprule
\textbf{Method} 
& \textbf{SC} 
& \textbf{PP} 
& \textbf{CP} 
& \textbf{D} 
& \textbf{OP} 
& \textbf{Average} \\
\midrule
Non-personalized 
& 1.10 & 1.15 & 1.18 & 1.06 & 1.12 & 1.12 \\

In-context Learning 
& 1.18 & 1.20 & 1.16 & 1.10 & 1.17 & 1.16 \\

RAG 
& 1.32 & 1.34 & 1.28 & 1.20 & 1.31 & 1.29 \\

SFT
& 2.08 & 1.95 & 1.93 & 1.44 & 1.88 & 1.86 \\

OPPU 
& \underline{2.18} & 2.05 & \underline{2.03} & 1.52 & \underline{2.00} & \underline{1.96} \\

AS
& 1.92 & 1.86 & 1.90 & 1.42 & 1.74 & 1.77 \\

Style Vector 
& 2.15 & \underline{2.08} & 1.98 & \underline{1.54} & 1.98 & 1.95 \\

Fints 
& 2.04 & 1.96 & 1.92 & 1.48 & 1.86 & 1.85 \\

GLASS 
& \textbf{2.35} & \textbf{2.14} & \textbf{2.20} & \textbf{1.60} & \textbf{2.16} & \textbf{2.09} \\
\bottomrule
\end{tabular}
}
\caption{
GPT-5 judge evaluation results on LaMP-4. 
SC, PP, CP, D, and OP denote Style Consistency, Preference and Persona Alignment, 
Content Personalization, Distinctiveness, and Overall Personalization, respectively. 
The best results are indicated in bold, and the second-best results are underlined.
}
\label{tab:gpt5_judge_lamp4}
\end{table}
\begin{table}[t]
\centering
\small
\setlength{\tabcolsep}{4pt}
\resizebox{\columnwidth}{!}{
\begin{tabular}{lcccccc}
\toprule
\textbf{Method} 
& \textbf{SC} 
& \textbf{PP} 
& \textbf{CP} 
& \textbf{D} 
& \textbf{OP} 
& \textbf{Average} \\
\midrule
Non-personalized
& 1.05 & 1.17 & 1.20 & 1.07 & 1.16 & 1.13 \\

In-context Learning 
& 1.35 & 1.32 & 1.27 & 1.14 & 1.27 & 1.27 \\

RAG
& 1.64 & 1.55 & 1.41 & 1.29 & 1.46 & 1.47 \\

SFT
& 2.34 & 2.13 & 2.22 & \underline{1.56} & 2.05 & 2.06 \\

OPPU 
& 2.28 & 2.10 & 2.14 & 1.54 & 2.00 & 2.01 \\

AS 
& 2.20 & 2.06 & 2.18 & 1.54 & 1.87 & 1.97 \\

Style Vector 
& \underline{2.31} & \textbf{2.15} & 2.11 & 1.55 & 1.99 & 2.02 \\

Fints 
& 2.30 & 2.08 & \underline{2.24} & 1.52 & \underline{2.08} & \underline{2.04} \\

GLASS 
& \textbf{2.50} & \underline{2.12} & \textbf{2.34} & \textbf{1.58} & \textbf{2.27} & \textbf{2.16} \\
\bottomrule
\end{tabular}
}
\caption{
GPT-5 judge evaluation results on LaMP-5.
SC, PP, CP, D, and OP denote Style Consistency, Preference and Persona Alignment,
Content Personalization, Distinctiveness, and Overall Personalization, respectively.
The best results are indicated in bold, and the second-best results are underlined.
}
\label{tab:gpt5_judge_lamp5}
\end{table}
\begin{table}[t]
\centering
\small
\setlength{\tabcolsep}{4pt}
\resizebox{\columnwidth}{!}{
\begin{tabular}{lcccccc}
\toprule
\textbf{Method} 
& \textbf{SC} 
& \textbf{PP} 
& \textbf{CP} 
& \textbf{D} 
& \textbf{OP} 
& \textbf{Average} \\
\midrule
Non-personalized
& 1.08 & 1.12 & 1.14 & 1.06 & 1.10 & 1.10 \\

In-context Learning 
& 1.20 & 1.18 & 1.16 & 1.10 & 1.17 & 1.16 \\

RAG
& 1.36 & 1.34 & 1.32 & 1.23 & 1.31 & 1.31 \\

SFT
& 1.95 & 1.86 & 1.88 & 1.42 & 1.80 & 1.78 \\

OPPU 
& 2.18 & 2.03 & 2.12 & 1.54 & 2.02 & 1.98 \\

AS
& 2.06 & 1.95 & 2.00 & 1.47 & 1.89 & 1.87 \\

Style Vector 
& \underline{2.22} & \underline{2.06} & 2.08 & 1.55 & 2.01 & 1.98 \\

Fints 
& 2.20 & 2.04 & \underline{2.28} & \underline{1.58} & \underline{2.10} & \underline{2.04} \\

GLASS 
& \textbf{2.40} & \textbf{2.12} & \textbf{2.30} & \textbf{1.64} & \textbf{2.20} & \textbf{2.13} \\
\bottomrule
\end{tabular}
}
\caption{
GPT-5 judge evaluation results on LongLaMP-3. 
SC, PP, CP, D, and OP denote Style Consistency, Preference and Persona Alignment, 
Content Personalization, Distinctiveness, and Overall Personalization, respectively. 
The best results are indicated in bold, and the second-best results are underlined.
}
\label{tab:gpt5_judge_longlamp3}
\end{table}
\begin{table}[t]
\centering
\small
\setlength{\tabcolsep}{4pt}
\resizebox{\columnwidth}{!}{
\begin{tabular}{lcccccc}
\toprule
\textbf{Method} 
& \textbf{SC} 
& \textbf{PP} 
& \textbf{CP} 
& \textbf{D} 
& \textbf{OP} 
& \textbf{Average} \\
\midrule
Non-personalized
& 1.05 & 1.10 & 1.12 & 1.04 & 1.08 & 1.08 \\

In-context Learning 
& 1.22 & 1.24 & 1.20 & 1.12 & 1.22 & 1.20 \\

RAG
& 1.60 & 1.55 & 1.50 & 1.28 & 1.52 & 1.49 \\

SFT
& 1.55 & 1.52 & 1.48 & 1.26 & 1.45 & 1.45 \\

OPPU 
& \underline{2.12} & \underline{2.00} & \underline{2.02} & \underline{1.50} & \underline{1.95} & \underline{1.92} \\

AS 
& 1.88 & 1.82 & 1.86 & 1.42 & 1.76 & 1.75 \\

Style Vector 
& 2.05 & 1.96 & 1.90 & 1.48 & 1.88 & 1.85 \\

Fints 
& 1.62 & 1.56 & 1.50 & 1.30 & 1.50 & 1.50 \\

GLASS 
& \textbf{2.20} & \textbf{2.06} & \textbf{2.08} & \textbf{1.56} & \textbf{2.05} & \textbf{1.99} \\
\bottomrule
\end{tabular}
}
\caption{
GPT-5 judge evaluation results on LongLaMP-4. 
SC, PP, CP, D, and OP denote Style Consistency, Preference and Persona Alignment, 
Content Personalization, Distinctiveness, and Overall Personalization, respectively. 
The best results are indicated in bold, and the second-best results are underlined.
}
\label{tab:gpt5_judge_longlamp4}
\end{table}

\subsection{Adaptive vs.\ Fixed Clustering}

\begin{table}[h]
  \centering
  \begin{tabular}{@{}lcc@{}}
    \toprule
    & \textbf{R-1} & \textbf{R-L} \\
    \midrule
    Fixed $K$            & 16.35 & 12.27 \\
    Adaptive $K$ (Ours)  & 17.42 & 12.88 \\
    \bottomrule
  \end{tabular}
  \caption{Effect of adaptive vs.\ fixed clustering on LongLaMP-3.}
  \label{tab:app_cluster}
\end{table}

Table~\ref{tab:app_cluster} compares adaptive clustering, which selects $K$ by maximizing the silhouette score over candidate values, against a fixed cluster assignment. Adaptive clustering yields consistent improvements, demonstrating that dynamically identifying interaction scenarios enables more precise local style modeling.

\subsection{SAE Feature Aggregation Strategy}

\begin{table}[h]
  \centering
  \begin{tabular}{@{}lcc@{}}
    \toprule
    \textbf{Strategy} & \textbf{R-1} & \textbf{R-L} \\
    \midrule
    Base (Dense) & 0.4403 & 0.3829 \\
    Mean (Ours)  & 0.4620 & 0.4064 \\
    Max          & 0.0308 & 0.0299 \\
    Last         & 0.0001 & 0.0001 \\
    \bottomrule
  \end{tabular}
  \caption{Comparison of SAE feature aggregation strategies on LaMP-5.}
  \label{tab:app_sae}
\end{table}

Table~\ref{tab:app_sae} compares four strategies for aggregating sparse features across tokens: (1) \textit{Base}, which bypasses the SAE and directly averages dense activations; (2) \textit{Mean}, which averages sparse activations over all response tokens (our default); (3) \textit{Max}, which takes element-wise maximum pooling; and (4) \textit{Last}, which uses only the final token representation. Mean aggregation substantially outperforms all alternatives, while Max and Last fail to produce meaningful style vectors. This validates that distributional averaging in the sparse domain is essential for capturing consistent stylistic patterns.

\subsection{Effect of Historical Data Size}

\begin{table}[h]
  \centering
  \begin{tabular}{@{}lccccc@{}}
    \toprule
    & $k$=1 & $k$=3 & $k$=5 & $k$=8 & $k$=10 \\
    \midrule
    R-1 & 18.02 & 17.88 & 18.72 & 18.21 & 18.51 \\
    R-L & 15.83 & 15.76 & 16.40 & 15.89 & 16.21 \\
    \bottomrule
  \end{tabular}
  \caption{Performance of GLASS under varying historical data sizes on LaMP-4.}
  \label{tab:app_data_size}
\end{table}

Table~\ref{tab:app_data_size} reports the effect of varying the number of historical instances used for style vector construction. GLASS maintains strong performance stability across all settings, with ROUGE-L ranging from 15.76 to 16.40. The results indicate that even with as few as one historical sample, GLASS produces competitive personalization, while $k=5$ provides the best trade-off between performance and computational efficiency.

\subsection{SAE Model Selection}
\label{app:sensitivity}

Table~\ref{tab:app_sae_model} evaluates GLASS with three different SAE architectures. Results show that switching across SAE models with 64$\times$ expansion (EleutherAI, THUKEG) produces comparable performance, confirming that GLASS is not tightly coupled to a specific SAE implementation. However, reducing the expansion factor to 8$\times$ (LlamaScope) leads to moderate degradation, indicating that sufficient latent dimensionality is necessary for the SAE to effectively disentangle stylistic features from semantic content.

\subsection{Offline Construction and Scalability}
\label{app:offline}

\begin{table}[h]
  \centering
  \resizebox{\columnwidth}{!}{%
  \begin{tabular}{@{}lccc@{}}
    \toprule
    \textbf{Method} & \textbf{Time (s/user)} & \textbf{Storage (MB/user)} & \textbf{Training-free} \\
    \midrule
    Personalized LoRA & $\sim$140 & $\sim$18 & $\times$ \\
    GLASS (Ours)      & $\sim$30  & $<$0.01 & $\checkmark$ \\
    \bottomrule
  \end{tabular}%
  }
  \caption{Offline construction cost comparison between GLASS and personalized PEFT methods.}
  \label{tab:app_offline}
\end{table}

We compare the offline construction cost of GLASS with personalized PEFT methods. GLASS requires only forward passes to construct the style vector bank, achieving approximately $4.7\times$ speedup in per-user construction time and over three orders of magnitude reduction in storage overhead compared to personalized LoRA. Furthermore, the precomputed style vectors exhibit strong temporal stability—on LaMP-5, performance drops by only 1.7\% per 100 new samples without any update—indicating that periodic refresh suffices for practical deployment.

\subsection{Steering Layers and Intervention Strength}
\label{app:layer_strength_fig}

\begin{figure*}[h]
  \centering
  \includegraphics[width=\linewidth]{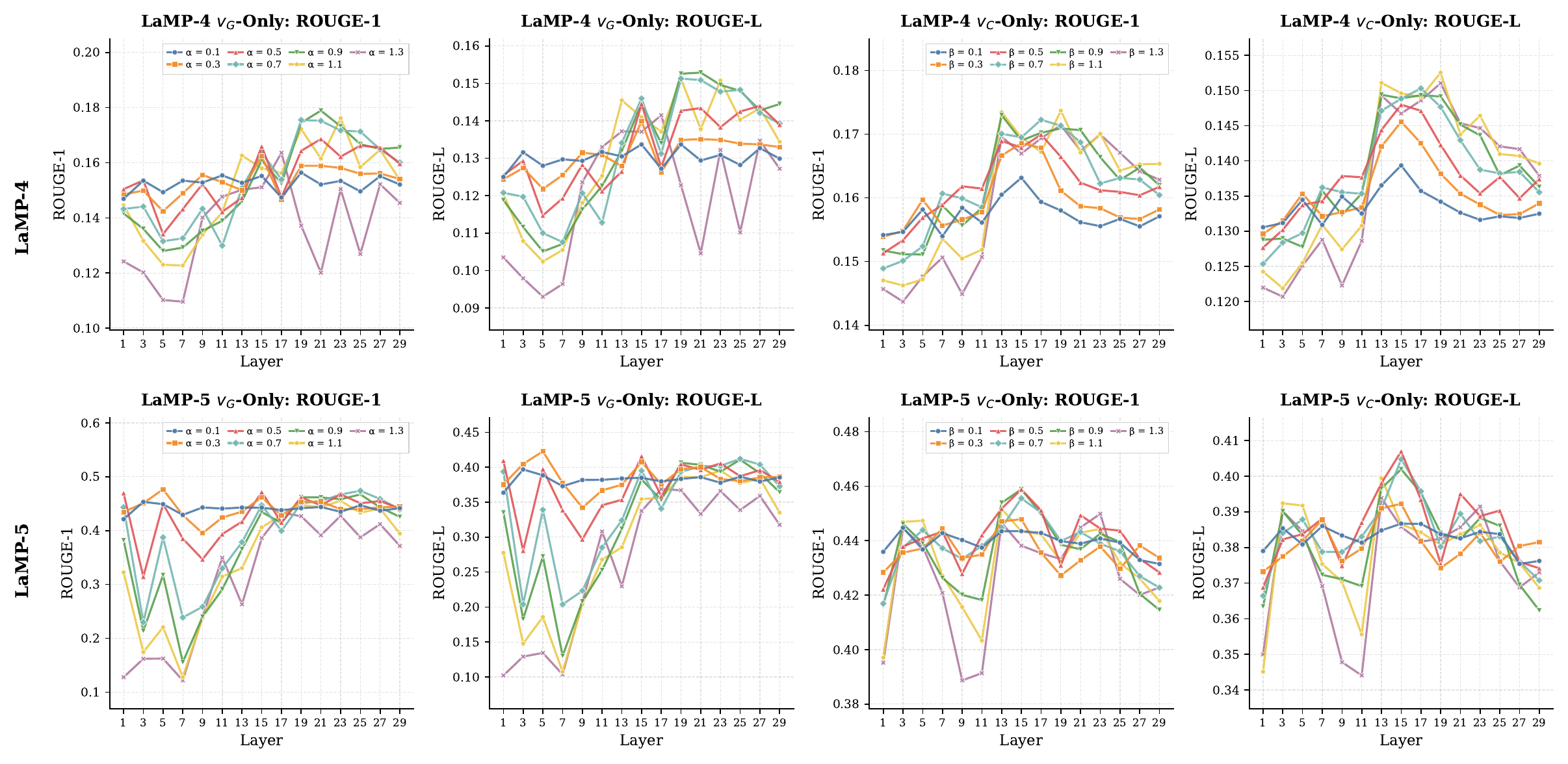}
  \caption{Performance comparison under different steering layers and intervention strength settings.}
  \label{fig:layer_strength}
\end{figure*}

Figure~\ref{fig:layer_strength} isolates the marginal effect of steering depth and intervention strength by evaluating global-only ($v_G$) and local-only ($v_C$) variants on LaMP-4 and LaMP-5. For each component, we sweep the intervention layer and the corresponding scaling coefficient, and report both ROUGE-1 and ROUGE-L. This diagnostic setting removes the interaction between the two streams, allowing us to examine where each style signal is most effective before considering their joint configuration.

\paragraph{Effect of intervention layers.}
Across both datasets, steering at very shallow layers is less reliable than steering at middle-to-upper layers. For the global vector $v_G$, the strongest regions generally appear after the middle of the network. On LaMP-4, the ROUGE-L curves rise noticeably around layers 19--23, especially under moderate coefficients. On LaMP-5, early-layer global steering is highly unstable, while layers after roughly 15 produce substantially more stable results. This pattern is consistent with the role of $v_G$: user-level stylistic preferences such as phrasing, compactness, and tone are better aligned with higher-level residual representations than with lower lexical features.

The local vector $v_C$ shows a related but more localized trend. Its best regions are concentrated around intermediate and middle-to-upper layers, where it can refine scenario-specific stylistic cues without fully overriding the task representation. In particular, the local-only curves on LaMP-4 improve sharply after the lower layers and peak around the middle-to-upper range, while the LaMP-5 curves show a narrower high-performing band. This suggests that $v_C$ is useful for context-dependent adjustment, but its effective layer range is more sensitive than that of $v_G$.

\paragraph{Effect of intervention strength.}
The results also show a clear non-monotonic relationship between intervention strength and performance. Very small coefficients often provide insufficient stylistic signal, whereas overly large coefficients increase variance across layers and can substantially hurt performance. This is especially visible for $v_G$ on LaMP-5, where strong early-layer steering causes large drops in both ROUGE-1 and ROUGE-L. Moderate strengths are therefore more reliable, because they allow the injected style direction to influence generation while preserving task-relevant semantics.

The two components differ in their sensitivity. The global vector is relatively robust once applied at suitable upper layers, which supports the claim that SAE-based sparse priors constrain the intervention toward stable user-level style dimensions. In contrast, the local vector exhibits sharper layer- and strength-dependent fluctuations, reflecting the fact that $v_C$ is extracted from contrastive local clusters and therefore carries more scenario-specific information. Over-amplifying this local signal can introduce instability even when the layer choice is reasonable.

\paragraph{Takeaways.}
These single-component sweeps indicate that effective personalization requires jointly calibrating where and how strongly to intervene. The best operating regions are not determined by strength alone: the same coefficient can help at one depth and degrade performance at another. The marginal peaks in Figure~\ref{fig:layer_strength} should therefore be viewed as diagnostic evidence rather than final hyperparameter choices; the full GLASS configuration is selected on validation data after accounting for interactions between $v_G$, $v_C$, and the SAE-primed construction procedure. Overall, the figure supports the main finding that global steering is better suited to upper-layer style modulation, while local steering provides complementary context-specific refinement when applied at compatible depths and moderate strengths.

\subsection{Pairwise Interaction Study}
\label{app:pairwise_fig}

\begin{figure*}[h]
  \centering
  \includegraphics[width=0.92\linewidth]{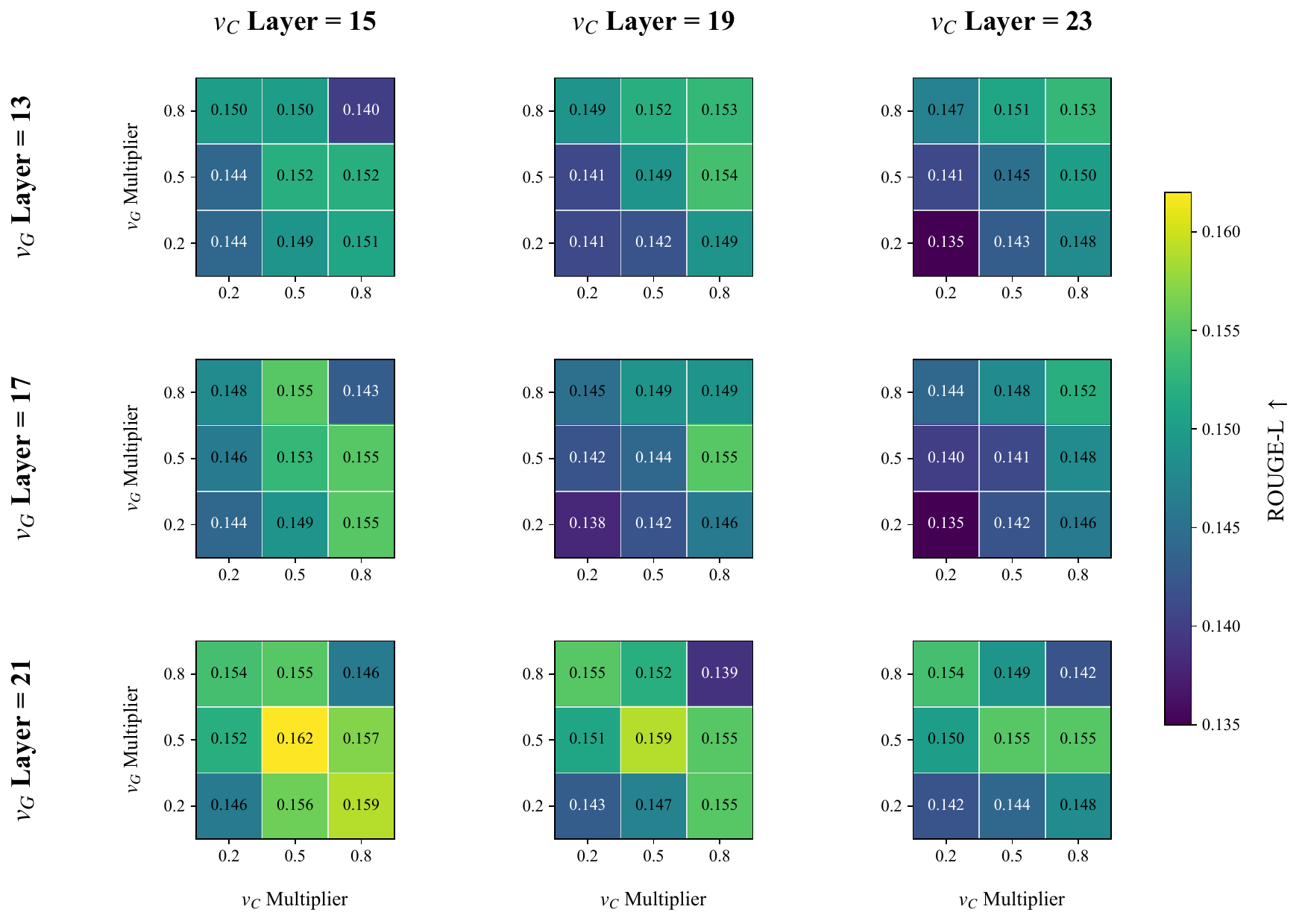}
  \caption{Pairwise interaction between the global style vector $v_G$ and the local contrastive vector $v_C$.}
  \label{fig:pairwise_vg_vc}
\end{figure*}

\begin{figure*}[h]
  \centering
  \includegraphics[width=0.92\linewidth]{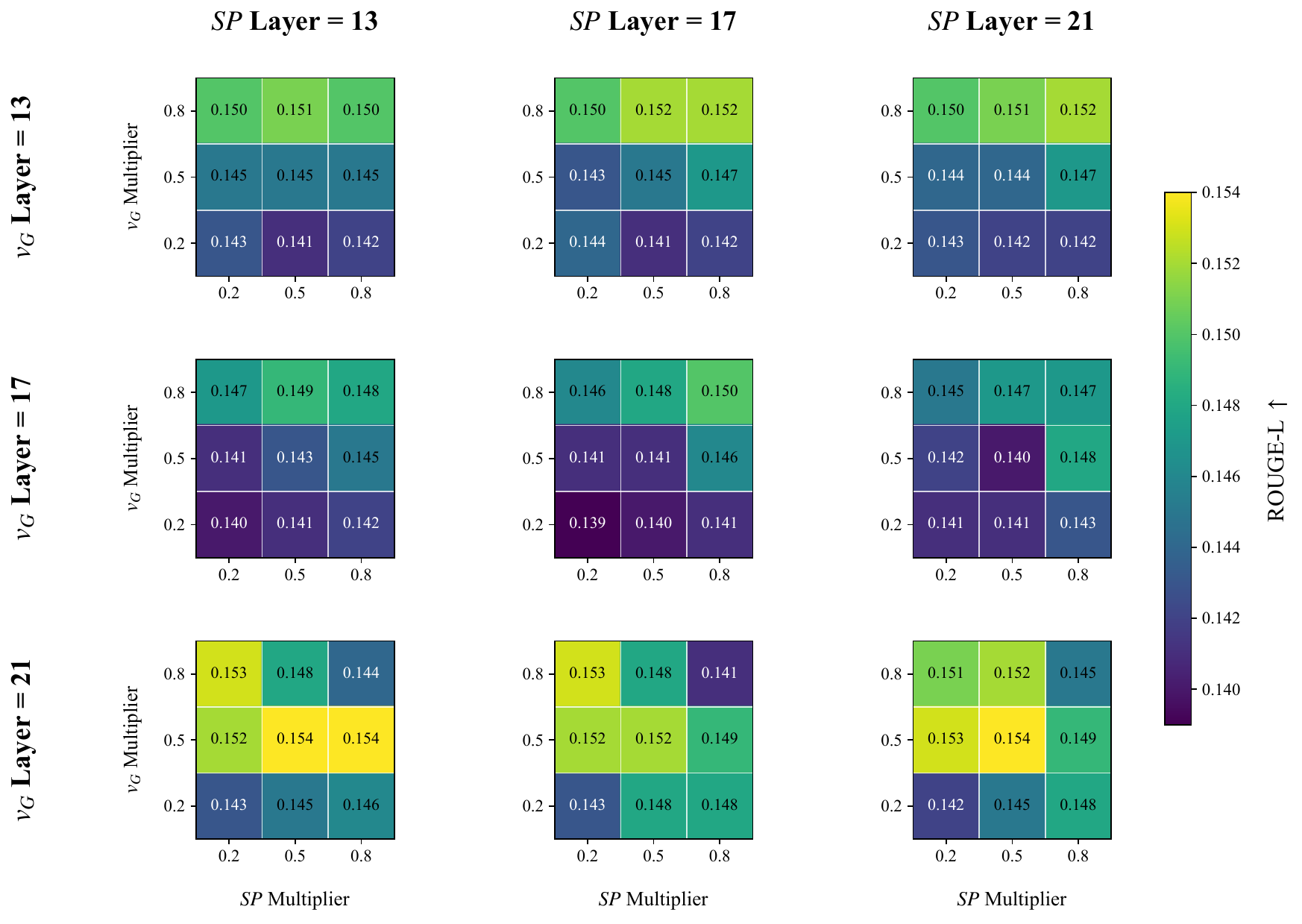}
  \caption{Pairwise interaction between the global style vector $v_G$ and the SAE-primed forward pass ($SP$).}
  \label{fig:pairwise_vg_sp}
\end{figure*}

Figures~\ref{fig:pairwise_vg_vc} and~\ref{fig:pairwise_vg_sp} further study pairwise interactions among GLASS components. Each heatmap reports ROUGE-L under a pair of layer and multiplier settings, with the remaining component fixed or averaged out to isolate the interaction of interest. Figure~\ref{fig:pairwise_vg_vc} varies the global style vector $v_G$ together with the local contrastive vector $v_C$, while Figure~\ref{fig:pairwise_vg_sp} varies $v_G$ together with the SAE-primed forward pass ($SP$). This fine-grained view complements Figure~\ref{fig:layer_strength} by showing that the optimal configuration depends on component compatibility rather than on independent single-component optima.

\paragraph{$v_G$--$v_C$ interaction.}
The $v_G$--$v_C$ heatmaps show the strongest complementarity. Performance generally improves when the $v_G$ layer is moved upward, with the best region appearing when the global style vector is applied at a higher layer and the local contrastive vector is applied at an intermediate layer. The highest observed ROUGE-L value appears around $v_G$ layer 21 and $v_C$ layer 15 with moderate multipliers, indicating that an upper-layer global style prior can provide a stable user-level scaffold while an earlier local vector refines the generation toward the current scenario.

The interaction is also sensitive to multiplier balance. Moderate settings for both components consistently outperform configurations where one component is too weak or both are too strong. When the $v_G$ multiplier is low, increasing the $v_C$ multiplier alone often yields only limited gains, suggesting that local steering benefits from a stable global prior. Conversely, when both multipliers are large or the $v_C$ layer is pushed too late, performance can drop, which indicates an over-steering effect: multiple strong perturbations may interfere with task-relevant representations rather than reinforce personalization. These patterns explain why the final GLASS configuration uses depth-aware dual-stream steering instead of simply maximizing each component independently.

\paragraph{$v_G$--$SP$ interaction.}
The $v_G$--$SP$ heatmaps exhibit a flatter and less synergistic landscape. Compared with $v_G$--$v_C$, varying the $SP$ layer changes performance within a narrower range, and the strongest patterns are still largely governed by the $v_G$ layer and multiplier. This supports the interpretation in the main text that the SAE-primed forward pass mainly improves offline local vector construction, rather than functioning as a strong independent inference-time steering signal. In particular, increasing $SP$ strength cannot compensate for an unsuitable $v_G$ configuration, while moderate $SP$ settings are sufficient when the global sparse prior is placed at an effective depth.

\paragraph{Takeaways.}
The pairwise study leads to three conclusions. First, GLASS benefits from depth-aligned steering: the global sparse prior $v_G$ is most useful at upper layers, while the local contrastive vector $v_C$ is most complementary at intermediate depths. Second, intervention strength is non-monotonic under joint steering; moderate multipliers provide the best balance between personalization and preservation of task semantics. Third, $v_C$ has stronger synergy with $v_G$ than $SP$, confirming that the primary inference-time benefit comes from combining global user-level style control with local context-dependent style refinement, whereas $SP$ mainly serves the offline construction of local vectors.

\section{Prompt Templates}
\label{app:prompts}

We use task-specific instruction prompts for all methods. For ICL and
retrieval-based baselines, the instruction is followed by user-history
demonstrations. GLASS uses the same task instruction but does not append
demonstrations; personalization is introduced through activation steering.
Fields in braces are substituted with the corresponding dataset fields.
Table~\ref{tab:prompt_templates} summarizes the prompt templates.

\begin{table*}[t]
  \centering
  \footnotesize
  \setlength{\tabcolsep}{4pt}
  \renewcommand{\arraystretch}{1.18}
  \begin{tabular}{@{}>{\raggedright\arraybackslash}p{0.22\textwidth}>{\raggedright\arraybackslash}p{0.73\textwidth}@{}}
    \toprule
    \textbf{Field} & \textbf{Content} \\
    \midrule
    \multicolumn{2}{@{}l@{}}{\textbf{Task-specific instruction prompts}} \\
    \midrule
    LaMP-4 & \texttt{Generate a headline for the following article: \{input\}.} \\
    LaMP-5 & \texttt{Generate a title for the following abstract of a paper: \{input\}.} \\
    LongLaMP-3 & \texttt{Generate the review text written by a reviewer who has a given an overall rating of "\{overall\}" for a product with description "\{description\}". The summary of the review text is "\{summary\}".} \\
    LongLaMP-4 & \texttt{Generate the content for a reddit post \{input\}.} \\
    \addlinespace[0.4em]
    \midrule
    \multicolumn{2}{@{}l@{}}{\textbf{Full prompt with user-history shots}} \\
    \midrule
    Few-shot/RAG baseline &
    \begin{minipage}[t]{0.72\textwidth}
    \ttfamily
    \#\#\# Instruction:\\
    \{task-specific instruction\}\\
    For your reference, here are the user's past QA pairs:\\
    Historical sample 0:\\
    Q: \{history\_input\_0\}\\
    A: \{history\_output\_0\}\\
    ...\\
    Historical sample k-1:\\
    Q: \{history\_input\_k-1\}\\
    A: \{history\_output\_k-1\}\\
    No explanation and additional words needed; just give your answer directly.\\[0.3em]
    Now it's your turn:\\
    \#\#\# \{output\_field\}:
    \end{minipage}
    \\
    \addlinespace[0.4em]
    \midrule
    \multicolumn{2}{@{}l@{}}{\textbf{Full prompt for GLASS}} \\
    \midrule
    Ours &
    \begin{minipage}[t]{0.72\textwidth}
    \ttfamily
    \#\#\# Instruction:\\
    \{task-specific instruction\}\\
    No explanation and additional words needed; just give your answer directly.\\[0.3em]
    Now it's your turn:\\
    \#\#\# \{output\_field\}:
    \end{minipage}
    \\
    \addlinespace[0.4em]
    \midrule
    Output field & \texttt{output\_field} is instantiated as \texttt{Headline} for LaMP-4, \texttt{Title} for LaMP-5, \texttt{Review} for LongLaMP-3, and \texttt{Post} for LongLaMP-4. \\
    \bottomrule
  \end{tabular}
  \caption{Prompt templates used in the experiments.}
  \label{tab:prompt_templates}
\end{table*}

\begin{table*}[t]
  \centering
  \footnotesize
  \setlength{\tabcolsep}{4pt}
  \renewcommand{\arraystretch}{1.12}
  \begin{tabular}{@{}>{\raggedright\arraybackslash}p{0.95\textwidth}@{}}
    \toprule
    \textbf{LLM-as-a-judge rubric prompt} \\
    \midrule
    \begin{minipage}[t]{0.93\textwidth}
    \ttfamily
    You are an impartial evaluator for personalized text generation. Your task is to judge whether a candidate output matches the target user's writing behavior.\\[0.2em]
    \#\#\# Task Instruction:\\
    \{task\_instruction\}\\[0.2em]
    \#\#\# Target User History:\\
    \{user\_history\}\\[0.2em]
    \#\#\# Test Input:\\
    \{x\_test\}\\[0.2em]
    \#\#\# Reference Response Written by the Target User:\\
    \{reference\_response\}\\[0.2em]
    \#\#\# Candidate Output:\\
    \{candidate\_output\}\\[0.2em]
    Evaluate the candidate output on a 1--5 integer scale for each criterion. Use the user history and reference response to infer the user's style, preferences, and writing habits, but do not score by lexical overlap alone.\\
    SC: Style Consistency, including tone, syntax, word choice, and formality.\\
    PP: Preference and Persona Alignment, including recurring preferences, organization, and voice.\\
    CP: Content Personalization, including content selection, emphasis, and framing.\\
    D: Distinctiveness from generic user-agnostic text.\\
    OP: Overall Personalization, i.e., the likelihood that the text was authored by the target user.\\
    Return only a JSON object with integer scores and short justifications for SC, PP, CP, D, and OP.
    \end{minipage}
    \\
    \bottomrule
  \end{tabular}
  \caption{Rubric prompt used for the GPT-5 judge evaluation.}
  \label{tab:judge_prompt}
\end{table*}



\section{Case Study}
\label{app:case_study}

We further inspect representative examples from LaMP and LongLaMP to illustrate how GLASS changes the generated text beyond semantic relevance. Tables~\ref{tab:case_study_short}, \ref{tab:case_study_longlamp3}, and \ref{tab:case_study_longlamp4} compare the reference output, the non-personalized baseline, and GLASS for the selected cases across LaMP-4/5 and LongLaMP-3/4.

\begin{table*}[t]
  \centering
  \footnotesize
  \setlength{\tabcolsep}{4pt}
  \renewcommand{\arraystretch}{1.18}
  \begin{tabular}{@{}>{\raggedright\arraybackslash}p{0.12\textwidth}>{\raggedright\arraybackslash}p{0.83\textwidth}@{}}
    \toprule
    \textbf{Field} & \textbf{Content} \\
    \midrule
    \multicolumn{2}{@{}l@{}}{\textbf{LaMP-4: Personalized News Headline Generation} \quad \textbf{ID:} \texttt{101800}} \\
    \midrule
    Reference & Instagram And Facebook Remove Posts Offering Abortion Pills \\
    Base & Facebook and Instagram began removing some of these posts, just as millions across the U.S. were searching for clarity around abortion access. \\
    GLASS & Facebook and Instagram began removing posts about abortion \\
    Reason & GLASS produces a more concise headline and removes the baseline's explanatory trailing clause, although it does not fully recover the Title Case convention. \\
    \addlinespace[0.6em]
    \midrule
    \multicolumn{2}{@{}l@{}}{\textbf{LaMP-5: Personalized Scholarly Title Generation} \quad \textbf{ID:} \texttt{111773}} \\
    \midrule
    Reference & Projected Subgradient Minimization Versus Superiorization. \\
    Base & The superiorization methodology for constrained minimization \\
    GLASS & Projected Subgradient Method for Constrained Minimization \\
    Reason & GLASS preserves the user's preferred terminology, especially \textit{Projected Subgradient}, and adopts a compact title-style noun phrase; the baseline uses a more generic formulation. \\
    \bottomrule
  \end{tabular}
  \caption{Case studies on short-form generation tasks (LaMP-4, LaMP-5) comparing references, non-personalized baseline outputs, and GLASS outputs.}
  \label{tab:case_study_short}
\end{table*}

\begin{table*}[t]
  \centering
  \footnotesize
  \setlength{\tabcolsep}{4pt}
  \renewcommand{\arraystretch}{1.18}
  \begin{tabular}{@{}>{\raggedright\arraybackslash}p{0.12\textwidth}>{\raggedright\arraybackslash}p{0.83\textwidth}@{}}
    \toprule
    \textbf{Field} & \textbf{Content} \\
    \midrule
    \multicolumn{2}{@{}l@{}}{\textbf{LongLaMP-3: Personalized Product Review Writing} \quad \textbf{ID:} \texttt{A5IXM5UERV8D2\#0}} \\
    \midrule
    Reference & I have been using this product for more than two weeks now. In that time, I've had an opportunity to evaluate both the quality of the product and the overall usability. In creating this review I've considered both the type of material used in making the product as well as the overall fit and finish. I've also considered how well the product meets my needs and whether there are any shortcomings or particular advantages in using this product. For this particular product, the type of material used in the manufacture and creation of the product seems to be of very good quality. One must consider that for any product of this type there is a balance between the type of material used and the cost that is to be passed on to the consumer. I think for this product that balance has been met fairly well and I believe that the product will meet the expectations of most buyers. With regard to the fit and finish of this particular product, I believe it to be about average for the products of similar make or style. That is to say that it is not of poor or moderate craftsmanship/finish nor is it something that I would rate as outstanding. The quality of the piece is such that I feel it is a good value for the money. \\
    Base & The type of material used in the manufacture and creation of the product seems to be of very good quality. One must consider that the product is made of a very durable material that will last for a long time. The product is also very easy to use and install. The product is also very easy to clean and maintain. \\
    GLASS & I have been using this product for a few months now and I have to say that I am very pleased with it. The type of material used in the manufacture and creation of the product seems to be of very good quality. One must consider that this is a product that is used in the bathroom and it is exposed to moisture and humidity. I have not had any issues with the product and it has been very reliable. I would recommend this product to anyone who is looking for a good quality towel bar. \\
    Reason & GLASS keeps the formal first-person review frame, reuses the deliberative phrase \textit{One must consider}, and avoids the baseline's repetitive loop. \\
    \bottomrule
  \end{tabular}
  \caption{Case study on the LongLaMP-3 product review writing task. Full long-form texts are included.}
  \label{tab:case_study_longlamp3}
\end{table*}

\begin{table*}[t]
  \centering
  \footnotesize
  \setlength{\tabcolsep}{4pt}
  \renewcommand{\arraystretch}{1.18}
  \begin{tabular}{@{}>{\raggedright\arraybackslash}p{0.12\textwidth}>{\raggedright\arraybackslash}p{0.83\textwidth}@{}}
    \toprule
    \textbf{Field} & \textbf{Content} \\
    \midrule
    \multicolumn{2}{@{}l@{}}{\textbf{LongLaMP-4: Personalized Topic Writing} \quad \textbf{ID:} \texttt{flupo42\#0}} \\
    \midrule
    Reference & Yesterday had this battle vs. AI - I inherited a confederated army that is mostly goblins. It has 5 night goblin archers, 2 night goblin melee units. 3 wolf rides, 1 Arrer Orc Boys, 2 Orc Boys and the rest Goblin Spearmen. So I am attacking an enemy Orc force. 7 of my units are hidden. The enemy waits for me to attack. I position my cavalry in a flanking position and the rest of my units on a hill. The enemy re-positions in response to my cavalry flank. Leaves one unit of their Boar Orc Boys a significant distance away from cover of the rest of their troops, facing my Wolf Riders. I see that the spacing between our armies and where that boar unit is leaves it exposed to a sudden attack by night goblins - if they all were to volley it from stealth it would either have to retreat back to mainline after taking massive damage, or try to charge them and be likely quickly killed by 2 night goblin melee units and focused fire of 5 archers. My wolf riders and the rest of my force should allow me to flank the rest of their line if it chooses to move to night goblins. So I begin the flanking maneuver with them - for like 6 minutes of game time everyone on the map except the goblins are still. I am not quite sure about the range of stealth, so I bring them in a wide crescent around the enemy Boar Orc Boys. In middle of that maneuver without any warning or notifications about stealth being uncovered, precisely at the worst possible time for me the enemy moves to attack my visible forces on the hill. My night goblins which comprise most of my missile troops are too far away to catch up - their flanking maneuver took them so far out they have a longer path back. I end up winning that battle with massive casualties (70\% of the army) because my goblins were left to hold the line vs enemy Savage Orcs without missile support and my Wolf Riders had to suicidally occupy 5 enemy Sage Orc Arrer Boys. \\
    Base & The AI knew the exact second when my hidden troops were out of position to support my visible troops. AI may not be able to target hidden units, but the tactical over-watch part of AI apparently knows exactly where they are. \\
    GLASS & I have previously tried to use the hide mechanic vs. AI a lot, and it always fails of backfires. However in those previous instances it relied on my troops hiding in trees or behind hills, and the AI wasn't obvious about 'finding' them - just seemed to always work out that it's flanks would go just wide enough to spot my ambushes. \\
    Reason & GLASS better matches the user's first-person tactical narration, hedging, quotation marks, and dash-connected reflective analysis, while the baseline is short and declarative. \\
    \bottomrule
  \end{tabular}
  \caption{Case study on the LongLaMP-4 topic writing task. Full long-form texts are included.}
  \label{tab:case_study_longlamp4}
\end{table*}

\section{AI Assistant Use}
We used ChatGPT to polish the language of the paper.

\end{document}